\documentclass[final,5p,times,twocolumn,authoryear]{elsarticle}
\usepackage{url,hyperref,lineno,microtype,subcaption}
\usepackage[onehalfspacing]{setspace}
\usepackage[thinc]{esdiff} % For nice derivative commands.
\usepackage{amssymb}
\usepackage{amsmath}
\usepackage{bm}
\usepackage{tikz}
\usepackage{adjustbox}
\usepackage{float} 
\usepackage{xcolor}
\definecolor{Tblue}{HTML}{1f77b4}
\definecolor{Torange}{HTML}{ff7f0e}
\definecolor{Tgreen}{HTML}{2ca02c}
\definecolor{Tred}{HTML}{d62728}
\definecolor{Tpurple}{HTML}{9467bd}
\definecolor{Tbrown}{HTML}{8c564b}

\usepackage{pgfplots}
\pgfplotsset{compat=1.17}

\makeatletter
\newenvironment{customlegend}[1][]{%
    \begingroup
    \pgfplots@init@cleared@structures
    \pgfplotsset{#1}%
}{%
    \pgfplots@createlegend
    \endgroup
}%

\def\addlegendimage{\pgfplots@addlegendimage}

\usepackage{pifont}
\usepackage[ruled,vlined]{algorithm2e}
\usepackage{algorithmic}
\usepackage{booktabs} % For \toprule \midrule and \bottomrule commands
\hypersetup{pdfauthor={Haakon Robinson}}

\newcommand{\numberofmodels}{10}
\usepackage{siunitx}

\modulolinenumbers[5]

\newcommand{\todo}[1]{}

\usepackage{soul}
\journal{Journal of \LaTeX\ Templates}
\begin{document}

\begin{frontmatter}
    \title{Physics guided neural networks for modelling of non-linear dynamics}% \\
    \author[adilsaddress]{Haakon Robinson} \ead{haakon.robinson@ntnu.no}
    \author[omersaddress]{Suraj Pawar} \ead{supawar@okstate.edu}
    \author[adilsaddress,adilSINTEFaddress]{Adil Rasheed\corref{mycorrespondingauthor}}
    \cortext[mycorrespondingauthor]{Adil Rasheed}
    \ead{adil.rasheed@ntnu.no}
    \author[omersaddress]{Omer San} \ead{osan@okstate.edu }

    \address[adilsaddress]{Department of Engineering Cybernetics, Norwegian University of Science and Technology}
    \address[omersaddress]{School of Mechanical and Aerospace Engineering, Oklahoma State University}
    \address[adilSINTEFaddress]{Mathematics and Cybernetics, SINTEF Digital}

    \begin{abstract}
        The success of the current wave of artificial intelligence can be partly attributed to deep neural networks, which have proven to be very effective in learning complex patterns from large datasets with minimal human intervention. However, it is difficult to train these models on complex dynamical systems from data alone due to their low data efficiency and sensitivity to hyperparameters and initialisation. This work demonstrates that injection of partially known information at an intermediate layer in a DNN can improve model accuracy, reduce model uncertainty, and yield improved convergence during the training. The value of these \textit{physics-guided neural networks} has been demonstrated by learning the dynamics of a wide variety of nonlinear dynamical systems represented by five well-known equations in nonlinear systems theory: the Lotka-Volterra, Duffing, Van der Pol, Lorenz, and Henon-Heiles systems.
    \end{abstract}

\end{frontmatter}
%\linenumbers

\section{Introduction}
\label{sec:introduction}
A dynamical system is a system whose state varies over time and obeys differential equations that involve time derivatives. The equations can either be analytically or numerically solved to predict the future state of the system. The evolution of the weather, progression of chemical reactions, the spread of diseases, and dynamics of vehicles can all be modelled as dynamical systems.

\begin{figure*}
    \centering
    \begin{subfigure}[b]{0.49\linewidth}
        {\includegraphics[width=\linewidth]{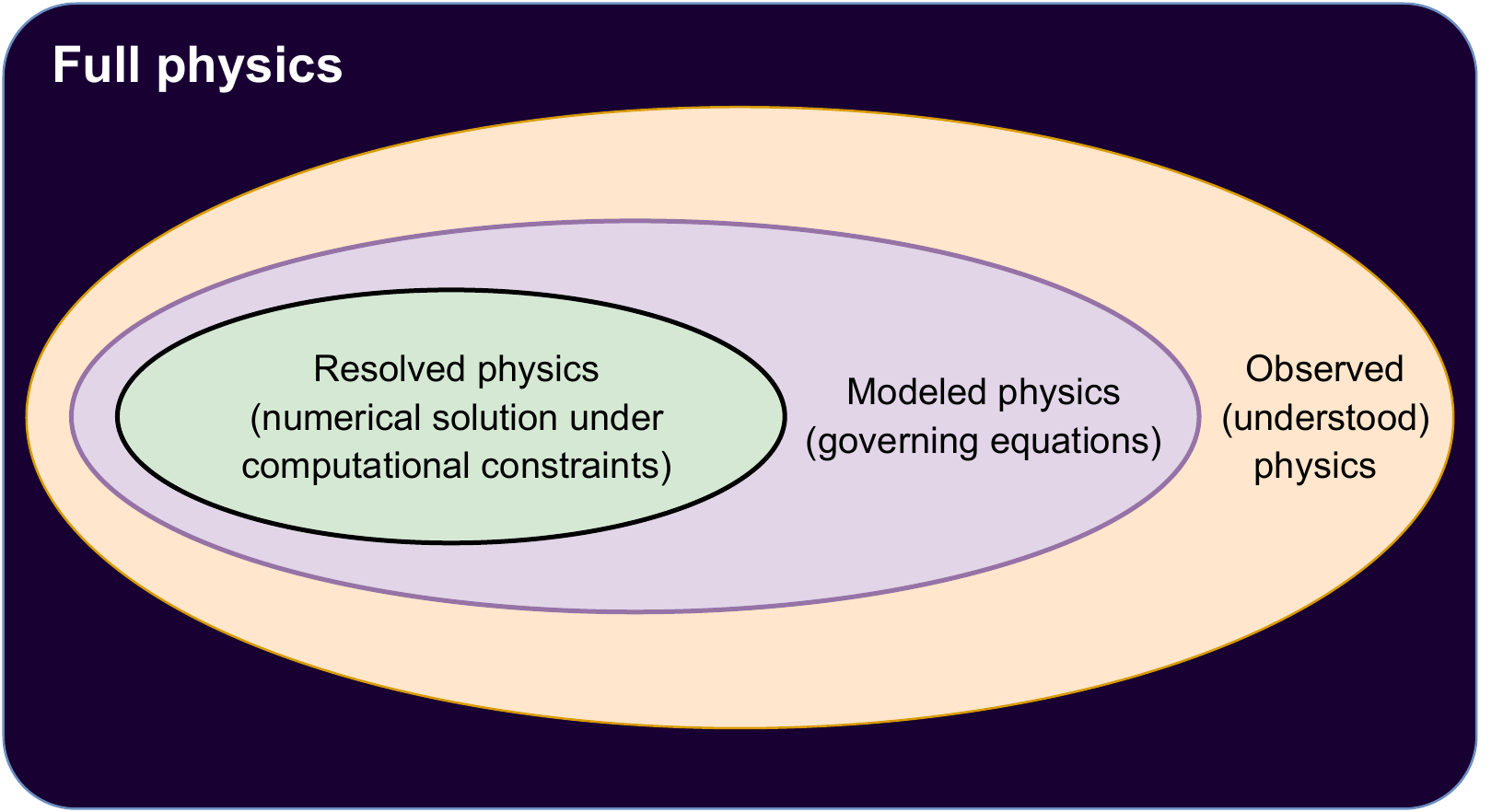}}
        \caption{PBM:Models represent a subset of full physics}
        \label{subfig:pbm}
    \end{subfigure}
    \begin{subfigure}[b]{0.49\linewidth}
        {\includegraphics[clip, trim={0.9cm 0.9cm 0.9cm 0.9cm}, width=\linewidth]{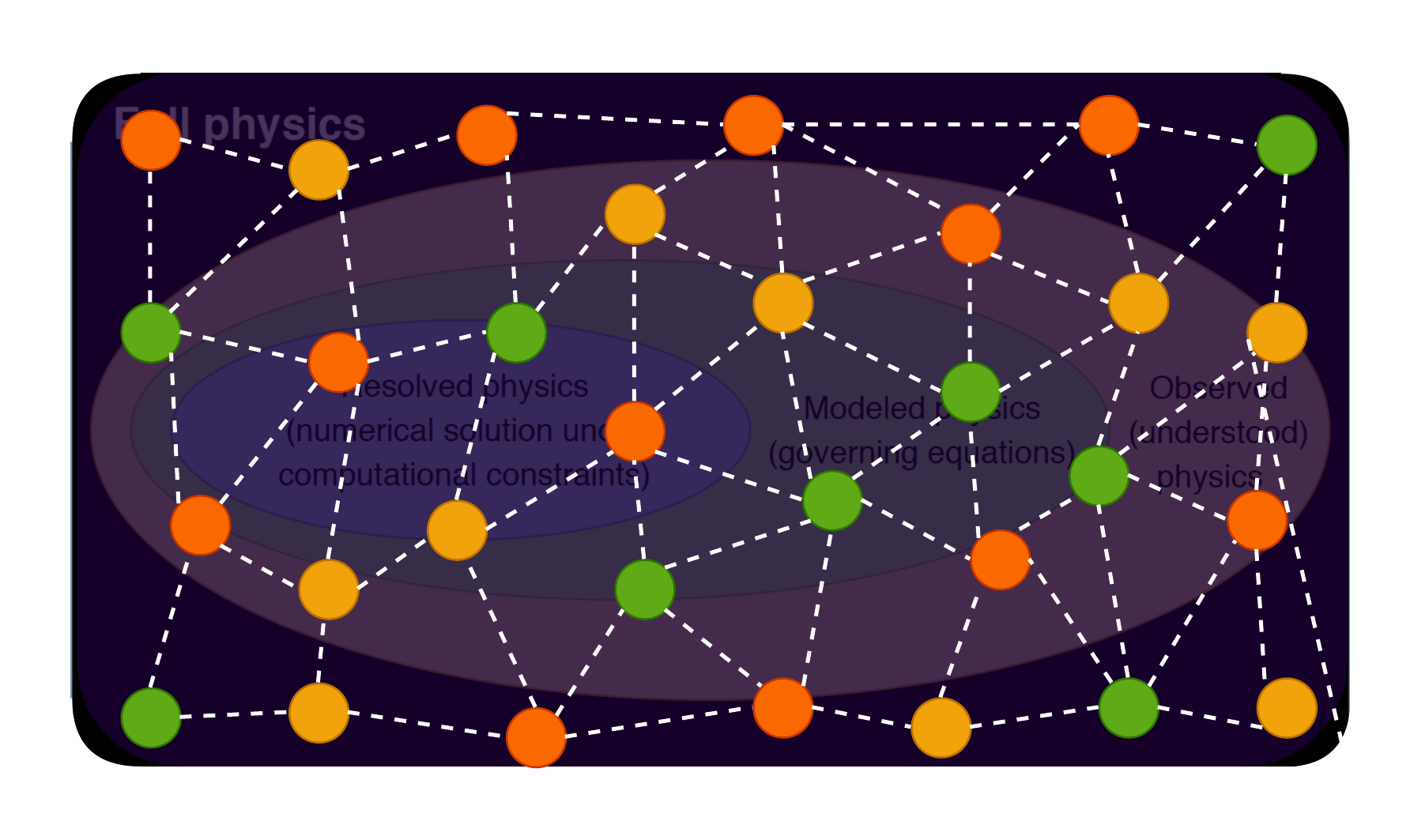}}
        \caption{DDM: Data can sample the full space}
        \label{subfig:ddm}
    \end{subfigure}
    \caption{Building a model naturally requires assumptions and simplifications. The result is that of the physics we can observe and understand, only a small part of the physics can be described using models, and even less can be numerically simulated. In contrast, large datasets can cover the full space, enabling general ML models to provide predictions in the absence of understanding or models.}
    \label{fig:full-physics}
\end{figure*}

Dynamical equations are typically derived from first principles using well understood physical laws. In this work, we call this approach physics-based modelling (PBM), and also use PBM to refer to the model itself (Figure \ref{subfig:pbm}). In developing a PBM, the observed physics are related to theory to produce equations, assumptions are made, and simplifications are sometimes imposed to make the solutions computationally tractable. PBMs typically have sound foundations from first principles, are interpretable, generalise well, and there exist robust theories for analysing properties such as stability and uncertainty. However, by relying on assumptions and simplifications, as well as our limited understanding of complex physical processes, we run the risk of not describing the desired phenomena with sufficient accuracy and certainty. Many PBMs do not account for unknown/unresolved physics, can be computationally expensive, do not adapt to new scenarios automatically and can be susceptible to numerical instabilities.

Data-driven modelling (DDM) (see Figure \ref{subfig:ddm}) is rapidly emerging as a tool that can address problems that resist traditional modelling methods, and some even controversially regard it as a full replacement for PBM \citep{karpathy_software_2017}.
These models can learn both known and unknown physics directly from data without prior knowledge of any physical laws, achieving good performance while maintaining computational efficiency for inference.
Deep neural networks (DNNs) in particular have enabled superhuman performance in tasks long considered impossible to solve for computers, such as the game of Go \citep{silver_mastering_2016}.
Works such as the protein-structure model proposed by \cite{senior_improved_2020} show that the use of DNNs has also begun to penetrate into scientific applications.
Within the realm of dynamical systems, DDMs have been demonstrated to learn dynamics directly from the data. For example, \cite{SAHA2021} use physics-incorporated convolutional recurrent neural networks for dynamical systems forecasting and source identification.

Despite these advantages, there remain some challenges before these models can find their way into high stake or safety-critical applications. They typically require vast amounts of data to train and can struggle to generalise to situations not represented well by the data. In contrast to established PBM methods, there is a lack of robust theory for the analysis of properties such as stability and robustness, and practitioners often have to fall back on empirical testing to assure the safety of their models.

To counter the issues associated with both the PBM and DDM, a new paradigm is emerging, which we call Hybrid Analysis and modelling (HAM). The HAM approach combines the generalisability, interpretability, robust foundation, and understanding of PBM with the accuracy, computational efficiency, and automatic pattern-identification capabilities of DDM. In their recent surveys, \cite{willard2020integrating} and \cite{san2021hybrid} provide comprehensive overviews of techniques for integrating DDM with PBM. Many of the hybridisation techniques fall into the following categories:
(i) Embedding PBMs inside NNs,
(ii) Model order reduction,
(iii) Physics-based regularisation terms,
(iv) Data-driven equation discovery,
(v) Error correction approaches, and
(vi) Sanity check mechanisms using PBMs.

In the following sections, we present related work and discuss the advantages and disadvantages of the approaches.

\subsection{Methods for embedding PBMs directly into NNs}
This is perhaps the most straightforward approach to hybridisation.
More advanced methods typically create a differentiable PBM that can be used as a layer in a NN.
An example of this is OptNet proposed by \cite{amos_optnet_2017}, a differentiable convex optimisation solver that can be used as a layer in a network.
In related work, \cite{avilabelbuteperes_end_2018} propose the differentiable physics engine, a rigid body simulator that can be embedded into a NN. They demonstrate that it is possible to learn a mapping from visual data to the positions and velocities of objects, which are then updated using the simulator. Similar ideas are used by \cite{yu_structural_2020} to simulate a structural dynamics problem by designing a hybrid recurrent NN (RNN) that contains an implicit numerical integrator. The advantage of these approaches is that they are usually quite data-efficient. A challenge is that these embedded PBMs are often iterative methods, making both inference and training more expensive.

\subsection{Model order reduction methods}
Reduced-order modelling (ROM) is a successful and widely adopted methodology \citep{quarteroni2014reduced}. A ROM method typically projects complex partial differential equations onto a lower dimensional space based on the singular value decomposition of the offline high fidelity simulation data. This yields a set of ordinary differential equations (ODEs) that can be efficiently solved \citep{ahmed2021closures}. ROMs have been used to accelerate high-fidelity numerical solvers by several orders of magnitude \citep{fonn2019dcp}. However, ROMs tend to become unstable in the presence of unknown/unresolved complex physics. To alleviate these problems, recent research has shown how unknown and hidden physics within a ROM framework can be accounted for using DNNs \citep{pawar2019aet,pawar2020ddr}. Despite these benefits, ROMs require full knowledge of the original equation before they can be applied.

\subsection{Physics-based regularisation terms}
By incorporating a PBM in the objective function, DDMs can be biased towards known physical laws during training.
A recent work \citep{raissi2019physics} that has seen a lot of interest is the physics-informed neural network (PINN), where a NN is used to represent the solution to a PDE and deviations from the equation at a sample of points are penalised by an additional loss term.
PINNs can be used to solve problems such as heat transfer, as was done by \cite{zobeiry_physics_2021} for parts in a manufacturing process.
The PINN approach has also been extended by \cite{arnold_state_2021} to allow for control in a state-space setting.
In related work, \cite{shen_physics_2021} create a model for classifying bearing health by training a NN on physics-based features and regularising the model using the output from a physics-based threshold model. These approaches require precise knowledge of the loss term. Complex regularization terms may impact the training process due to the increased cost of computing the loss function.

\subsection{Data-driven equation discovery}
Sparse regression based on $l_1$ regularization and symbolic regression based on gene expression programming have been shown to be very effective in discovering hidden or partially known physics directly from data. Notable work using this approach can be found in \cite{Champion22445} and \cite{vaddireddy2020fes}. One of the limitations of this class of method is that, in the case of sparse regression, additional features are required to be handcrafted, while in the case of symbolic regression, the resulting models can be often unstable and prone to overfitting.

\subsection{Error correction and sanity check mechanisms}
Corrective Source Term Approach (CoSTA) is a method proposed by \cite{blakseth2022deep} that explicitly addresses the problem of unknown physics. This is done by augmenting the governing equations of a PBM with a DNN-generated corrective source term that takes into account the remaining unknown/neglected physics. One added benefit of the CoSTA approach is that the physical laws can be used to keep a sanity check on the predictions of the DNN used, i.e. checking conservation laws. A similar approach has also been used to model unresolved physics in turbulent flows \citep{maulik2019sgm,pawar2020apa}. However, even these approaches assume a specific structure for at least the known part of the equation.

\subsection{Proposal: Physics-guided neural networks}
From the previous discussion, it is clear that almost all the HAM approaches discussed above require information about the structure of the equation representing the physics, which is not always available. We often have a very simplistic understanding of the physics. For example, we can have some understanding of the diurnal variation of solar radiation but not about its influences on atmospheric flow. To exploit even such a small amount of knowledge, \cite{pawar2021pgml} proposed a physics guided machine learning (PGML) approach. The basic idea behind the PGML approach is to inject partial knowledge into one of the layers within a DNN to guide the training process. The partial knowledge can, for example, come from a simplistic model or an empirical law \citep{pawar2021model,pawar2022multi}.

This paper extends the PGML concept to modelling nonlinear dynamical systems. Since we limit the model space to neural networks, we call the approach physics-guided neural network (PGNN). Through a series of experiments involving a variety of equations representing nonlinear dynamical systems like Lotka-Volterra, Duffing, Van der Pol, Lorenz, and Henon-Heiles equations, we attempt to answer the following questions:
\begin{itemize}
    \item What are the effects of knowledge injection on the training convergence?
    \item How does the accuracy/performance change with the choice of injection layer?
    \item Is there any correlation between model uncertainty and knowledge injection?
\end{itemize}
A brief background and rationale for the proposed method is given in Section \ref{sec:rationale}. Section \ref{sec:setup} details the selected dynamical systems that are considered in our study. Finally, the results are discussed in Section \ref{sec:resultsanddiscussion}, and conclusions and recommendation for future work made in Section \ref{sec:conclusionandfuturework}.

\section{Physics-guided neural networks}
\label{sec:rationale}
\begin{figure}
    \centering
    \includegraphics[width=\linewidth]{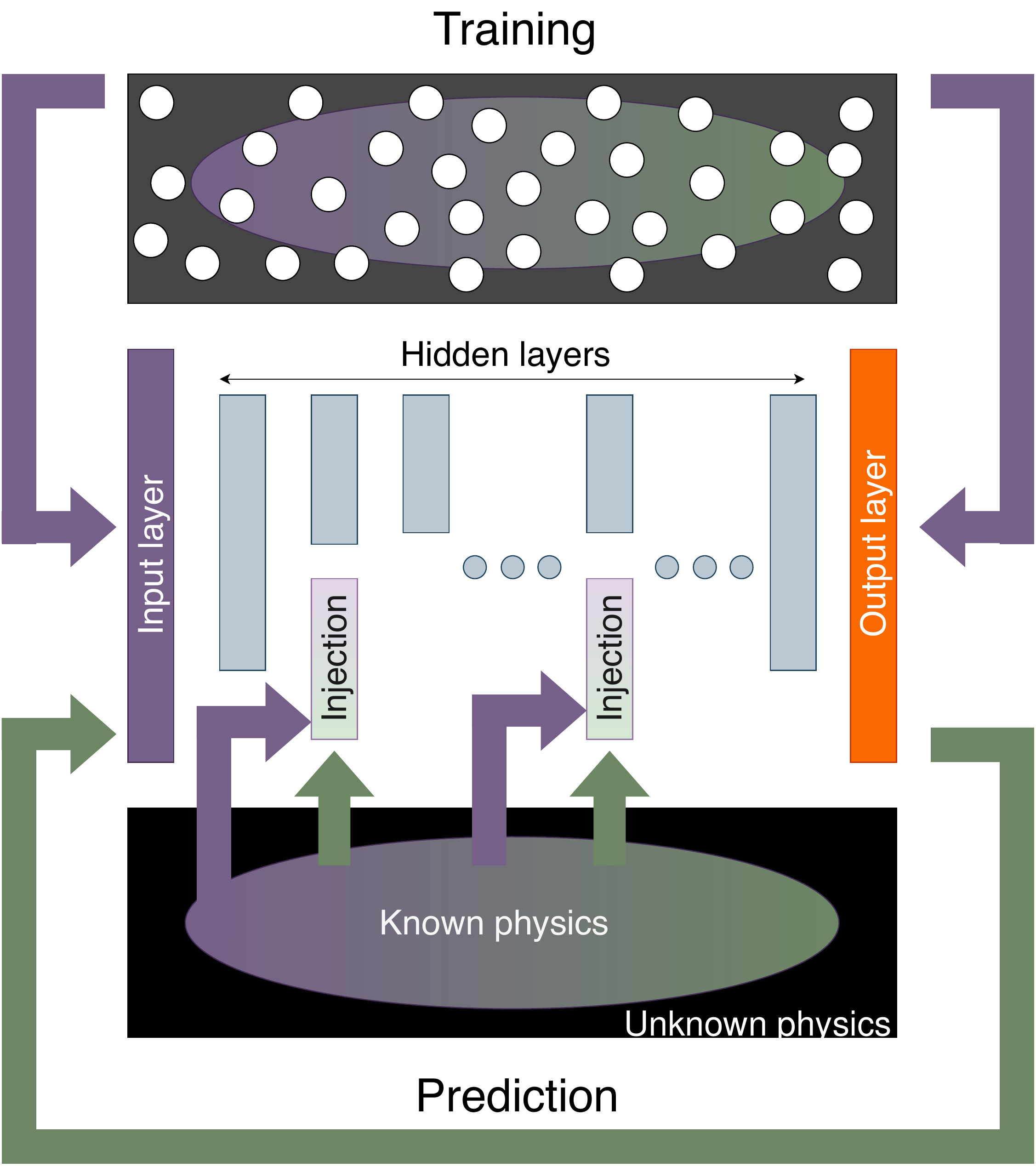}
    \caption{PGNN framework: The purple arrows correspond to the training phase while the green arrows correspond to the prediction phase. Assume that the data is represented by the white circles.)
    }
    \label{fig:pgml}
\end{figure}
The basic idea behind PGNN is to generalise the Principal Component Regression (PCR). In PCR, instead of regressing the dependent variable on the explanatory variables directly, the latent variables derived from the explanatory variables after the application of PCA are used as regressors. By replacing the high dimensional explanatory variables (containing redundancy) with much lower dimensional latent variables as the input to the regression model, one can significantly reduce the complexity of the regressors and make them more robust. However, there are two major problems associated with PCR. Firstly, the latent variables computed using PCA can only be a linear combination of the explanatory variables. Secondly, the regression task is decoupled from the latent variable computation. 

In the PGNN approach, both the computation of the latent variables and the regression are combined within a neural network framework with bottleneck layers representing the latent variable layers. Additionally, the latent variables can be supplemented with additional features (partial knowledge) to improve the accuracy and reduce the uncertainty of the trained PGNN model. If the additional features were used in combination with the explanatory features as input to the DNN, chances are high that they would get corrupted during the training process.

\newcommand{\state}[1]{\mathbf{x}_{#1}}
\newcommand{\dyn}[1]{\mathbf{f}\left(#1\right)}
\newcommand{\learneddyn}[1]{\hat{\mathbf{f}}\left(#1\right)}

We now present the rationale behind the PGNN approach to modelling nonlinear dynamics. Deep learning has been recently used in many studies to model the spatio-temporal dynamics of high-dimensional systems \citep{vlachas2018data,pathak2018model,pawar2019deep}. Given some dynamical system
$\dot{\state{}} = \dyn{\state{}}$,
a NN can be trained directly on the mapping $\dyn{\cdot}$ by sampling repeatedly from the system. After training, the network can then be numerically integrated in order to perform predictions on the future states of the system, e.g. by computing the forward Euler step $\state{k+1} = \state{k} + h\,\learneddyn{\state{k}}$. Consider a dataset generated by a more general dynamical system:

\begin{equation}
    {\cal L} \mathbf{x} = f(g(\mathbf{x}),h(\mathbf{x}))
    \label{eq:generalequation}
\end{equation}

where ${\cal L}$ is a linear differential operator, and $f(\cdot)$, $g(\cdot)$ and $h(\cdot)$ are functions of the state. Now, the following scenarios can arise:
\begin{enumerate}
    \item Equation \eqref{eq:generalequation} is fully known meaning that the operator $\cal L$,  and the functions $f(.)$, $g(.)$ and $h(.)$ are precisely known
    \item The operator $\cal L$ is known but one or two of the functions $f(\cdot)$, $g(\cdot)$ and $h(\cdot)$ are unknown
    \item The operator $\cal L$ is known, but the functions $f(\cdot)$, $g(\cdot)$ and $h(\cdot)$ are all unknown
    \item The operator $\cal L$ and the functions $f(\cdot)$, $g(\cdot)$ and $h(\cdot)$ are all unknown
\end{enumerate}

In the first scenario, one can employ a purely PBM approach based on the known equations. The only advantage of DDM over PBM is possibly superior computational performance, which may enable real-time applications.
In the second and third scenarios, while the problem can in theory be solved entirely using DDM, it would be unwise to ignore the known part completely. Incorporating them into the DDM may simplify the learning task as well as improve generalisation. In the fourth scenario, PBM is impossible, and it is necessary to model the process entirely from data. 

Assume now that only $h(x)$ is known in Equation \eqref{eq:generalequation}. Then \eqref{eq:generalequation} can be learned using a PGNN with an $h(x)$ injected at hidden layer of the neural network, as shown in Figure \ref{fig:pgml}. By stacking known features into an intermediate layer, they can be utilised more effectively. The significance of which layer is used for injection is unknown.

Here, we briefly explain the NN and PGNN architecture. A neural network is designed using several layers consisting of a predefined number of neurons. Each neuron has a weighted connection to all neurons in the previous layer, and a bias term. This is represented as an affine transformation as shown below
\begin{equation}
    \mathbf{z}^l = \mathbf{W}^l \boldsymbol{\chi}^{l-1} + \mathbf{b}^l,
\end{equation}
where $\boldsymbol{\chi}^{l-1}$ is the output of the $(l-1)^{\text{th}}$ layer, $\mathbf{W}^l$ is the matrix of weights representing the incoming connection strengths the $l^{\text{th}}$ layer, and $\mathbf{b}^l$ is the bias vector. For notational simplicity we define $\boldsymbol{\chi}^0 = \mathbf{x}$. The transformed input is then passed through a node's activation function $\zeta$, which is some nonlinear function. The introduction of this nonlinearity prevents the chain of affine transformations from simplifying and allows the neural network to learn highly complex relations between the input and output. The output of the $l^{\text{th}}$ layer can be written as
\begin{equation}
    \boldsymbol{\chi}^l = \zeta(\mathbf{z}^l),\;\;\;\;\; \boldsymbol{\chi}^0 = \mathbf{x}
\end{equation}
where $\zeta$ is the activation function. Some possible choices are the ReLU, tanh, and sigmoid activation functions. We refer the reader to \mbox{\cite{goodfellow_deep_2016}} for a more complete overview. If there are $L$ layers between the input and the output in a neural network, then the output of the neural network can be represented mathematically as follows

\begin{equation}
    \dot{\mathbf{x}} =\zeta_L(\mathbf{W}^L,\mathbf{b}^{L},\dots,\zeta_2(\mathbf{W}^2,\mathbf{b}^{2},\zeta_1(\mathbf{W}^1,\mathbf{b}^{1},\mathbf{x})))
\end{equation}
where $\mathbf{x}$ and $\dot{\mathbf{x}}$ are the independent and dependent variables of the system, respectively. The above equation can also be written as

\begin{equation}
    \dot{\mathbf{x}} = \zeta_{L}(\cdot ; \boldsymbol{\Theta}_{L}) \circ \cdots \circ \zeta_{2}(\cdot ; \boldsymbol{\Theta}_{2}) \circ \zeta_{1}(\mathbf{x} ; \boldsymbol{\Theta}_{1})
\end{equation}

where $\boldsymbol{\Theta}$ represents the weights and biases of the corresponding layer of the neural network. For the PGNN framework, the information from the known part of the system is injected into an intermediate layer of the neural network as follows

\begin{equation}
    \dot{\mathbf{x}} = \zeta_{L}(\cdot ; \boldsymbol{\Theta}_{L}) \circ \cdots
    \circ \underbrace{\mathcal{C}(\zeta_{i}(\cdot ; \boldsymbol{\Theta}_{i}),h(\mathbf{x}))}_{\text{Known function injection}} \circ
    \cdots \circ \zeta_{1}(\mathbf{x} ; \boldsymbol{\Theta}_{1}),
\end{equation}

where $\mathcal{C}(\cdot, \cdot)$ represents the concatenation operation and the known information about the system, i.e., $h(\mathbf{x})$ is injected at $i$th layer. However, the choice of this layer is significant, and there is currently no way to know a priori which layer will yield the best results. In this paper, we investigate this by providing knowledge injections at each layer.

\section{Methodology}
\label{sec:setup}
To test the applicability of the PGNN approach, experiments were performed on five nonlinear dynamical systems. For each system, suitable injection terms were identified. The same NN architecture (3 hidden layers) was used in all cases to reduce the number of experiments. The functions were then injected with the following configurations: no injection, injection in the first layer, second layer, and third layer. Then, for each injection configuration, an ensemble of \numberofmodels\ models was trained on the data. This was done in order to estimate the model uncertainty.

\subsection{Choice of equations}
The systems were selected to cover a wide range of nonlinear phenomena including periodic and aperiodic solutions, limit cycles, and chaos. These properties are shown in Table \ref{tab:props}.

\newcommand{\ymark}[0]{\ding{51}}
\newcommand{\nmark}[0]{ }

\begin{table}
    \centering
    \caption{Possible types of solutions of the chosen nonlinear systems for the chosen parameters.}
    \begin{tabular}{lccc}
        \hline
        \textbf{System} & \textbf{Periodic} & \textbf{Limit cycle} & \textbf{Chaotic} \\ \hline
        Lotka-Volterra  & \ymark            & \nmark               & \nmark           \\
        Duffing         & \nmark            & \nmark               & \ymark           \\
        Van der Pol     & \ymark            & \ymark               & \nmark           \\
        Lorenz          & \nmark            & \nmark               & \ymark           \\
        Henon-Heiles    & \ymark            & \nmark               & \ymark           \\ \hline
    \end{tabular}
    \label{tab:props}
\end{table}

\subsubsection{Lotka-Volterra}
The Lotka–Volterra equations are often used to describe the interactions of a population of predators $x$ and a population of prey $y$:

\begin{align}\label{eq:lotka-volterra2}
    \begin{split}
        \dot{x} &= \alpha x - \beta xy, \\
        \dot{y} &= \delta xy - \gamma y,
    \end{split}
\end{align}
where $\dot{y}$ and $\dot{x}$ represent the instantaneous growth rates of the two populations due to predation, overpopulation, and starvation. The solutions are periodic; as the prey population $x$ grows, the predators $y$ can eat more and reproduce. This leads to a decline in $x$, causing $y$ to drop as the predators starve. The two variables thus appear as similar waves, with $y$ lagging behind $x$. In this paper, the values $\alpha=0.1, \beta=0.05, \delta=0.1, \gamma=1.1$ were used.

The Lotka-Volterra equations, and predator-prey models in general, remain of theoretical and practical interest today. Such systems have been successfully used to model ecological communities \citep{bunin_ecological_2017}, infections \citep{ghanbari_mathematical_2020}, and economic growth cycles \citep{goodwin_growth_1982,veneziani_structural_2006,harvie_dynamical_2007}. We will attempt to inject the nonlinear term $xy$, as it appears in both equations.

\subsubsection{Duffing}
\label{sec:duffing-bg}
The Duffing equation is a non-linear second-order differential equation that describes an oscillator with complex, sometimes chaotic behaviour. The Duffing equation was originally the result of Georg Duffing's systematic study of nonlinear oscillations \citep{hamel_georg_1921}. Interest in the equation was later revived with the advent of chaos theory. Since then, the system has come to be regarded as one of the prototype systems in chaos theory \citep{strogatz_nonlinear_2015}, and related equations continue to find applications today, e.g. to describe the rolling of ships \citep{wawrzynski_bistability_2018}. The equation is

\begin{align}\label{eq:duffing-raw}
    %\ddot{x}  &= 3\mathrm{cos}(0.4 t) -  \dot{x} - 0.5 x - x^3 - 0.1 x^5.
    \ddot{x} & = \gamma \mathrm{cos}(\omega t) -  \delta \dot{x} - \alpha x - \beta x^3 %- 0.1 x^5.
\end{align}
where $x(t)$ is the displacement at time $t$ and the term $\gamma \mathrm{cos}(\omega t)$ represents a sinusoidal driving force. The cubic term describes an asymmetry in the restoring force of a spring that softens or stiffen as it is stretched. The parameters used in this work are $\delta=1,\alpha=0.5,\beta=1,\gamma=3$ and $\omega=0.4$. Note that this is a time varying system depending on $t$. This is challenging to model using a neural network as-is, as the input $t$ is unbounded. Instead, we reparametrise the system as follows:
\begin{equation}\label{eq:duffing}
    \begin{aligned}
        \dot{x}      & = y                                              \\
        \dot{y}      & = \gamma \psi -  \delta y - \alpha x - \beta x^3 \\
        \dot{\psi}   & = -\omega \theta                                 \\
        \dot{\theta} & = \omega \psi                                    \\
        \psi(0)      & = 1,\;\; \theta(0) = 0
    \end{aligned}
\end{equation}
This enables us to treat the system as if it were time-invariant. Note that from the ML perspective this is equivalent to feature engineering, as we provide the features $\mathrm{cos}(\omega t)$ and $\mathrm{sin}(\omega t)$ as additional inputs to the model. For knowledge injection, we use the $x^3$ term, and we also reuse the $\mathrm{cos}(\omega t)$ term to see if providing redundant features has any effect.

\subsubsection{Van der Pol oscillator}
The Van der Pol equation was discovered by \cite{vanderpol_theory_1960} while studying triode vibrations. It describes a nonlinear oscillator that approaches a limit cycle over time. Systems like this are immensely useful in a variety of fields. For example, coupled Van der Pol systems have been used by \cite{rompala_dynamics_2007}  to model biological circadian rhythms, and by \cite{lucero_modeling_2013} to model the asymmetries in vocal folds. \cite{kuiate_autonomous_2018} have even applied a variant of the system to encrypt images in real-time.
The Van der Pol oscillator can be written as:
\begin{equation}\label{eq_vdp}
    \begin{aligned}
        \dot{x} & = y                \\
        \dot{y} & = \mu(1-x^2)y + x.
    \end{aligned}
\end{equation}
where $x(t)$ is the displacement, and $\mu=3$ is a scalar that controls the effects of the nonlinear damping term. In this work, we set $\mu=3$.

The system tends to a stable limit cycle for all initial conditions. As $x$ approaches the maximum amplitude of the oscillation, $\dot{x}$ increases. When reaching the maximum, $\dot{x}$ rapidly switches sign and $x$ begins to decrease slowly, building up speed in the same way as it approaches the minimum. When the equation is forced with an additional sinusoidal term it can exhibit chaos, however this is not done in this work. The $x^2y$ is fairly complex, and we select this for knowledge injection.

\subsubsection{Lorenz system}
The Lorenz system was originally developed to describe atmospheric convection by \cite{lorenz_deterministic_1963}, but later became one of the most well-studied systems in chaos theory, and is often credited with the explosion of interest in the subject \citep{strogatz_nonlinear_2015}. The Lorenz equations have since been studied in connection with real physical phenomena, such as unstable spiking in lasers \citep{haken_analogy_1975} and turbulence \citep{ruelle_lorenz_1976}.

The system has the following form:
\begin{align}\label{eq:Lorentz}
    \begin{split}
        \dot{x} &= \sigma(y-x)\\
        \dot{y} &= x(\rho - z) - y\\
        \dot{z} &= xy - \beta z
    \end{split}
\end{align}

For $\rho<1$, the origin is globally stable. When $\rho>1$, the system has three fixed points: $(0,0,0)$, and $(\pm \sqrt{(\beta(\rho-1)},\pm \sqrt{(\beta(\rho-1)},\rho-1)$, the latter of which we call $C_+$ and $C_-$. For $\rho>\frac{\sigma(\sigma+\beta+3)}{\sigma-\beta-1}$ the solutions of the system become non-periodic and chaotic where almost all initial states will converge to an invariant fractal set called the Lorenz attractor \citep{viswanath_fractal_2004}. Here we use $\sigma=10$, $\rho=28$, and $\beta=8/3$, the values originally used by Lorenz. The terms $xy$ and $xz$ are natural candidates for injection. In this work we only test $xy$.

\subsubsection{Henon--Heiles}
\cite{henon_applicability_1964} originally developed these equations to study the movement of a star around a galactic centre while restricted to a plane. The system is still used to study the escape dynamics of orbits \citep{zotos_comparing_2015}. It is governed by the following Hamiltonian:
\begin{equation}
    H = \frac{1}{2} (\dot{x}^2 + \dot{y}^2) + \frac{1}{2}(x^2+y^2) + x^2y - \frac{y^3}{3}
\end{equation}
This can be reformulated as a set of ODEs:
\begin{equation}
    \begin{aligned}
        \ddot{x} & = -x - 2\lambda x y         \\
        \ddot{y} & = -y - 2\lambda (x^2 - y^2)
    \end{aligned}
\end{equation}
In this work we set $\lambda=1$.
The solution set features a large number of periodic orbits, chaotic orbits, and escape trajectories when the energy of the system is sufficiently high \citep{zotos_classifying_2014}. The escape sets exhibit a rich fractal structure which adds additional complexity to the system behaviour \citep{zotos_overview_2017}. This system has several features that can be injected. In this work, we try $xy$ and $y^2$, with $x^2$ omitted because it appears in the equation similarly to $y^2$.

\subsection{Data generation and pre-processing}
\label{sec:data-generation}
For each of the five dynamical systems, we generated training data for the neural networks, and a test set to judge if the trained model can generalize to previously unseen states. This was done by manually choosing a set of initial conditions $\mathrm{x}_0$ and generating the corresponding trajectories using the RKF45 solver with adaptive timestepping until a final time $T$. We used the implementation in the SciPy software stack {\citep{2020SciPy_NMeth}}, which is based on the Dormand-Prince pair of formulas {\citep{dormand_family_1980}}. The resulting data was then interpolated to generate a regular timeseries with timestep $h$. The time derivative at each data point was estimated as the forward difference $(f(x^+) - f(x))/h$. The datasets can then be described as a list of pairs ${\cal D} = \{(\mathbf{x}_k,\mathbf{y}_k)\}$, where $\mathbf{x}_k$ and $\mathbf{y}_k$ are the $k$th state and time derivative respectively. A validation set was constructed by reserving $20\%$ of the data. The models are never trained on the validation set, but we monitor their performance on this data in order to measure their performance on unseen data. The initial conditions and other parameters that were used for each system are provided in Table {\ref{tab:dataset-params}}. The test trajectory was generated from the last initial condition for each system, as discussed in Section {\ref{sec:resultsanddiscussion}}.
 
\begin{table*}
    \centering
    \caption{Parameters and initial conditions used to generate the datasets. The training set was constructed by simulating each system with the initial conditions shown below using RKF45 with adaptive timestep until time $T$, and then estimating the pairs $(\mathbf{x}(t),\dot{\mathbf{x}}(t))$ at regular time intervals of length $h$. 20\% of these pairs were reserved for the validation set. The test set was generated from a different initial condition, as shown below.}
    \begin{adjustbox}{max width=\linewidth}
        \begin{tabular}{cccll}
            \hline Model   & Timestep $h$ (s) & Final time $T$ (s) & Initial conditions (IC) for training and validation & IC for testing \\ \hline
            Lotka-Volterra & $0.05$       & $200$                     & $(2,1), (10,1), (12,1), (15,1), (20,1), (22,1), (25,1) $ & $(5,1)$\\
            Duffing        & $0.05$       & $200$                     & $(1,1), (0,1), (-1,1), (1,-1), (0,-1), (-1,-1) $ & $(1,0.5)$ \\
            Van der Pol    & $0.005$      & $20$                      & $ (0,6), (0,-2), (-1,2), (1,-4), (0,0.1), (1,3), (-2,5) $ & $(2,-5)$\\
            Lorenz         & $0.005$      & $25$                      & $(1,1,1), (5,1,1), (1,5,1), (1,1,5), (-5,1,1), (1,-5,1)  $ & $(1,1,-5)$\\
            Henon-Heiles   & $0.05$       & $100$                     & $ (0.1,0.5,0,0), (0.3,0.4,0,0), (-0.35,0.4,0,0), (0.3,-0.1,0,0) $ & $(-0.325,0.4,0,0)$ \\
            \hline
        \end{tabular}
    \end{adjustbox}
    \label{tab:dataset-params}
\end{table*}

\subsection{Neural network architectures and training}
\label{subsec:nn-architectures}
The same network architecture was used in all cases to allow for a better comparison. The networks were given 3 hidden layers with 32, 64, and 32 neurons respectively. The injection was performing by concatenating the injection term to the selected hidden layer. Each model ensemble consisted of 10 neural networks. This was found to yield decent uncertainty estimates. The models were implemented in Tensorflow \citep{martinabadi_tensorflow_2015} and trained using the ADAM optimiser \citep{kingma_adam_2014} with default parameters. The models were trained on batches of 32 samples at a time (this number is known as the batch size) for a total of 100 epochs. An epoch is defined as the number of batch iterations after which the model will have trained on all data within the training set. We describe each batch as a set of indices ${\cal B} \subset {\cal N}$ that correspond to data in $\cal D$. 

Since this is a regression problem, the mean-squared error (MSE) was utilised as a loss function. The loss for the training batch $\cal B$ is then

\begin{equation}
    L_{MSE}({\cal B} ; \boldsymbol{\theta}) = \sum_{k\in{\cal B}} \Vert \mathbf{y}_k - \hat{f}(\mathbf{x}_k ; \boldsymbol{\theta}) \Vert
\end{equation}
where $(\mathbf{x}_k, \mathbf{y}_k)$ is the $k$th pair in the dataset $\cal D$, as described in Section \ref{sec:data-generation}. Regularization methods such as weight decay are usually used during training to prevent overfitting. We did not encounter any overfitting issues, and therefore we do not apply any regularisation to reduce the number of comparisons.

\subsection{Model evaluation}
\label{sec:model_evaluation}
It was found that simply reporting the MSE on the test set did not clearly show how the models performed. Therefore, we chose to report the model performance as the MSE between a rolling forecast and the test trajectory, which we refer to as the rolling forecast MSE (RFMSE). During the rolling forecast stage, the initial condition for the first time step is provided. This information is used to predict the forecast state at the next time step using a forward Euler step, which is then used to predict the next time step until the final time step is reached. The timesteps shown in Table \ref{tab:dataset-params} were used. The resulting trajectories of the model ensemble were then compared to the true trajectory of the system. We believe that reporting the RFMSE more accurately reflects the actual use case of these models and makes it easier to qualitatively see how knowledge injection can affect the predictive accuracy and model uncertainty within each model class.

\section{Results and discussion}
\label{sec:resultsanddiscussion}
In this section, we report the performance of the ensembles in terms of their training/validation loss, as well as the RFMSE on the test trajectory (see Section \ref{sec:model_evaluation}).
The model uncertainty within each model class is shown using 95\% confidence bounds around the average predictions. We also report the mean training and validation loss for each ensemble. The loss signals of all models were smoothed using an exponential moving average filter using a weight of $0.2$ before being averaged. This was done to improve the clarity of the plots, and allows us to compare overall trends between ensembles as well as the stability of the training.
First, an overview of the results is presented, and then we provide a more detailed look at the best performing injection term within each model class.

\subsection{Overview of results}
\begin{figure}
    \centering
    \includegraphics[width=\linewidth]{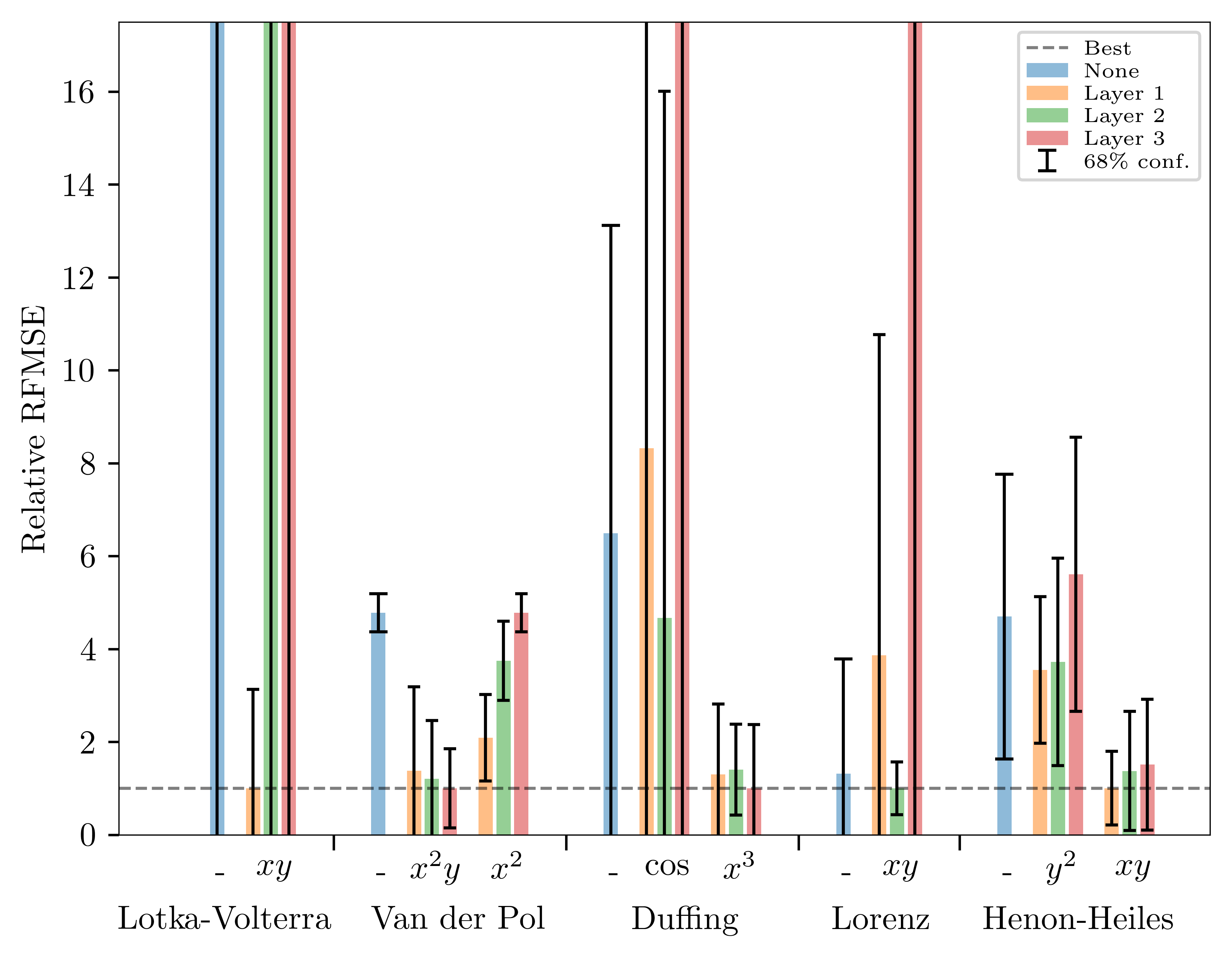}
    \caption{The relative RFMSE for all model ensembles across different systems. The values for each system have been divided by the minimum RFMSE in each group for better comparison, such that the top performers for each system have a value of 1. Note that in some cases the rolling forecasts have diverged. Because of this, we do not compute the RFMSE on the full trajectory. Instead we use the final times $[25\si{\second}, 70\si{\second},2.5\si{\second},2.5\si{\second},15\si{\second}]$ for each system respectively. The 68\% confidence intervals shown here were chosen to improve clarity while still allowing for a comparison between models.}
    \label{fig:testing_mse2}
\end{figure}

The RFMSE for each model class is visualised in Figure \ref{fig:testing_mse2}. Note that the data has been normalised due to the different scales of each test set, such that a value of 1 represents the top performer for each system. Additionally, because the predictions of the models blow up in some cases, we compute the RFMSE on a shorter time interval: $[25\si{\second}, 70\si{\second},2.5\si{\second},2.5\si{\second},15\si{\second}]$ for the Lotka-Volterra, Duffing, Lorenz, Van der Pol, and Henon-Heiles systems respectively. For all systems, the best performing ensemble on average was a PGNN, often by a significant margin.
The choice of injection layer appears to be a significant factor, although at this stage the data shows no conclusive pattern. This is surprising, as all of the nonlinear terms show up as additive terms in the equations, and there does not appear to be a good reason for the difference. Methods such as layer-wise relevance propagation \mbox{\citep{montavon_layer_2019}} could be adopted to interpret the impact of the choice of layer for knowledge injection and we consider it as part of our future work.

\subsection{Lotka Volterra system with \texorpdfstring{$xy$}{xy} injection}
\begin{figure}
    \centering
    \begin{adjustbox}{max width=0.75\linewidth}
        \begin{tikzpicture}
            \begin{customlegend}[legend columns=4,legend style={draw=none, column sep=1ex},legend entries={No injection, $xy$ layer 1, $xy$ layer 2, $xy$ layer 3}]
                \addlegendimage{black,sharp plot}
                \addlegendimage{Tblue,sharp plot}
                \addlegendimage{Torange,sharp plot}
                \addlegendimage{Tgreen,sharp plot}
            \end{customlegend}
        \end{tikzpicture}
    \end{adjustbox}

    \begin{subfigure}[b]{0.475\linewidth}
        \includegraphics[width=\linewidth]{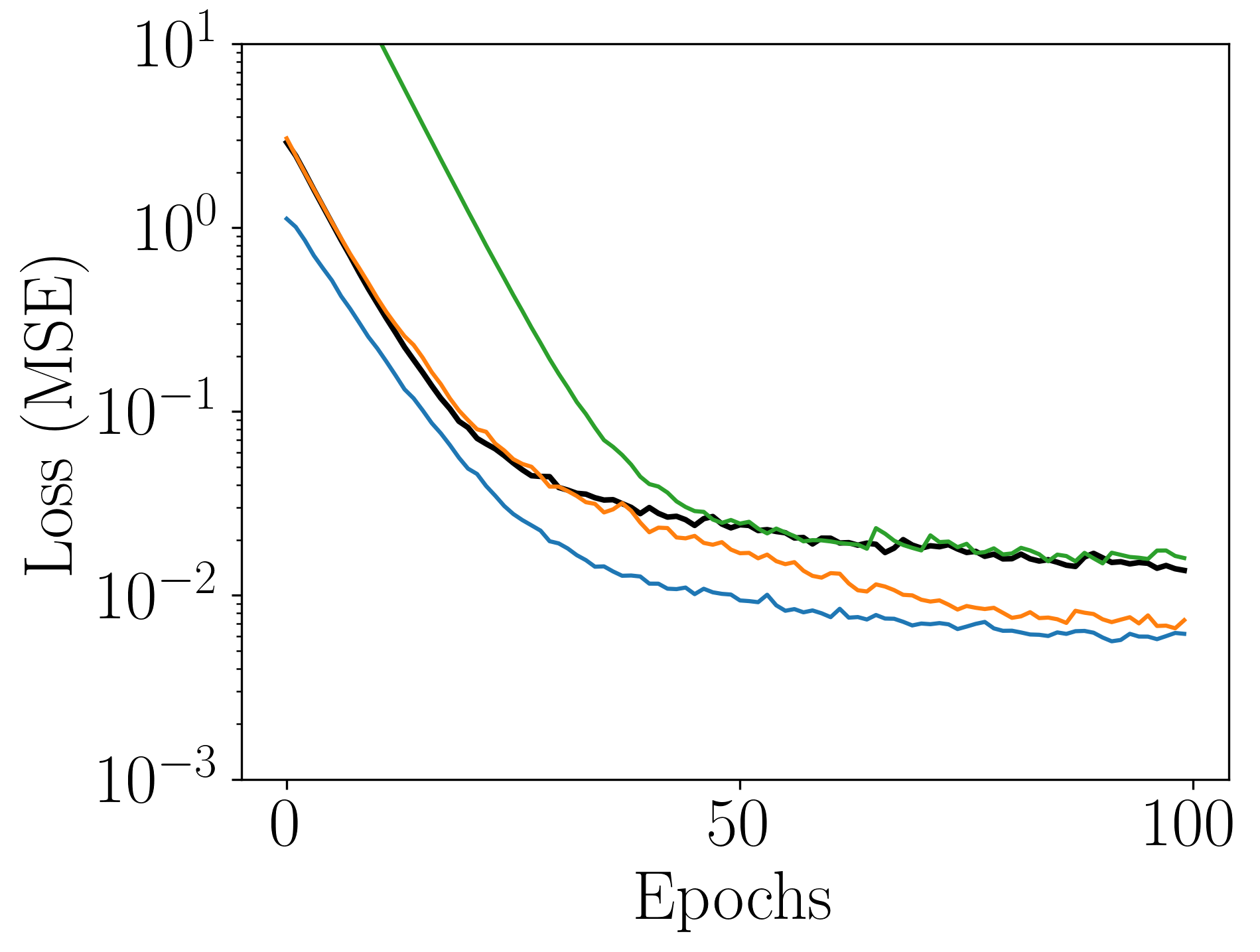}
        \caption{Training loss}
    \end{subfigure}
    \begin{subfigure}[b]{0.475\linewidth}
        \includegraphics[width=\linewidth]{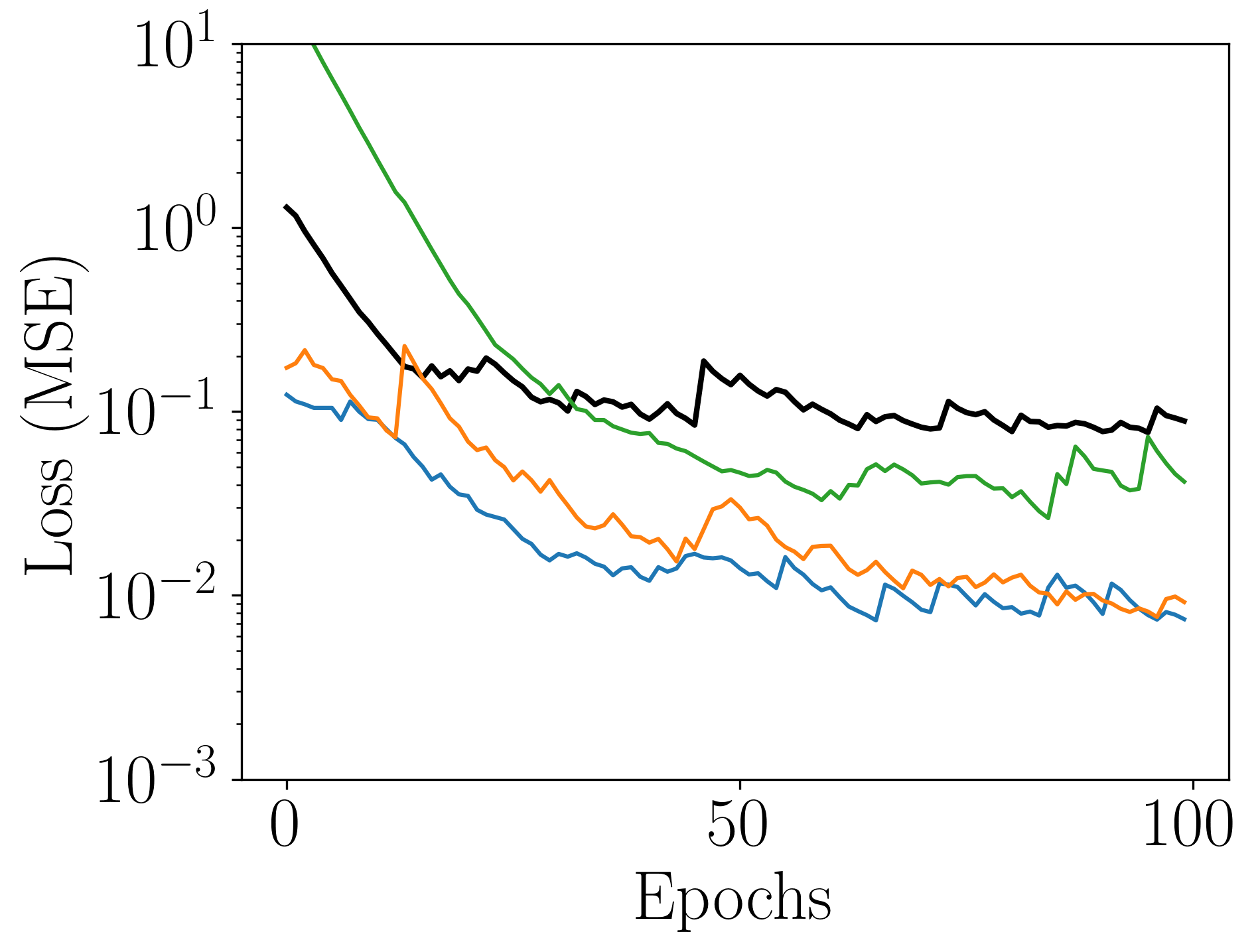}
        \caption{Validation loss}
    \end{subfigure}
    \caption{Comparison of the training and validation loss for the Lotka-Volterra system for different injection configurations.}
    \label{fig:lotkavolterra_training_loss_comparison}
\end{figure}

\begin{figure}
    \centering
    \begin{adjustbox}{max width=0.5\linewidth}
        \begin{tikzpicture}
            \begin{customlegend}[legend columns=3,legend style={draw=none, column sep=1ex},legend entries={Truth,Prediction,95\% conf.,$x$,$y$}]
                \addlegendimage{black, sharp plot}
                \addlegendimage{black, dashed,sharp plot}
                \addlegendimage{black!20, fill=black!20, area legend}
                \addlegendimage{Tblue,sharp plot}
                \addlegendimage{Torange,sharp plot}
            \end{customlegend}
        \end{tikzpicture}
    \end{adjustbox}

    \begin{subfigure}[b]{0.475\linewidth}
        {\includegraphics[width=\linewidth]{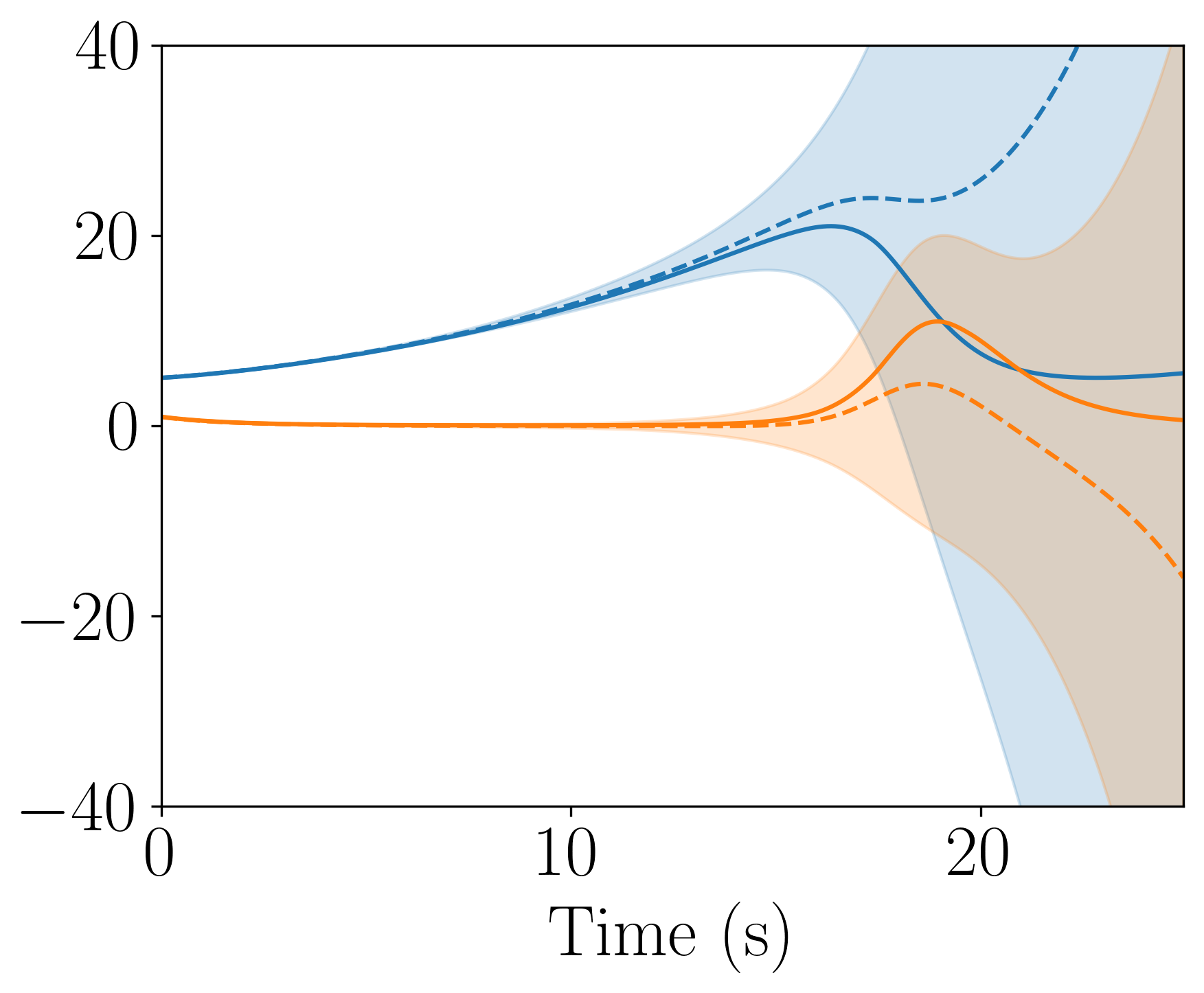}}
        \caption{No injection}
    \end{subfigure}
    \begin{subfigure}[b]{0.475\linewidth}
        \includegraphics[width=\linewidth]{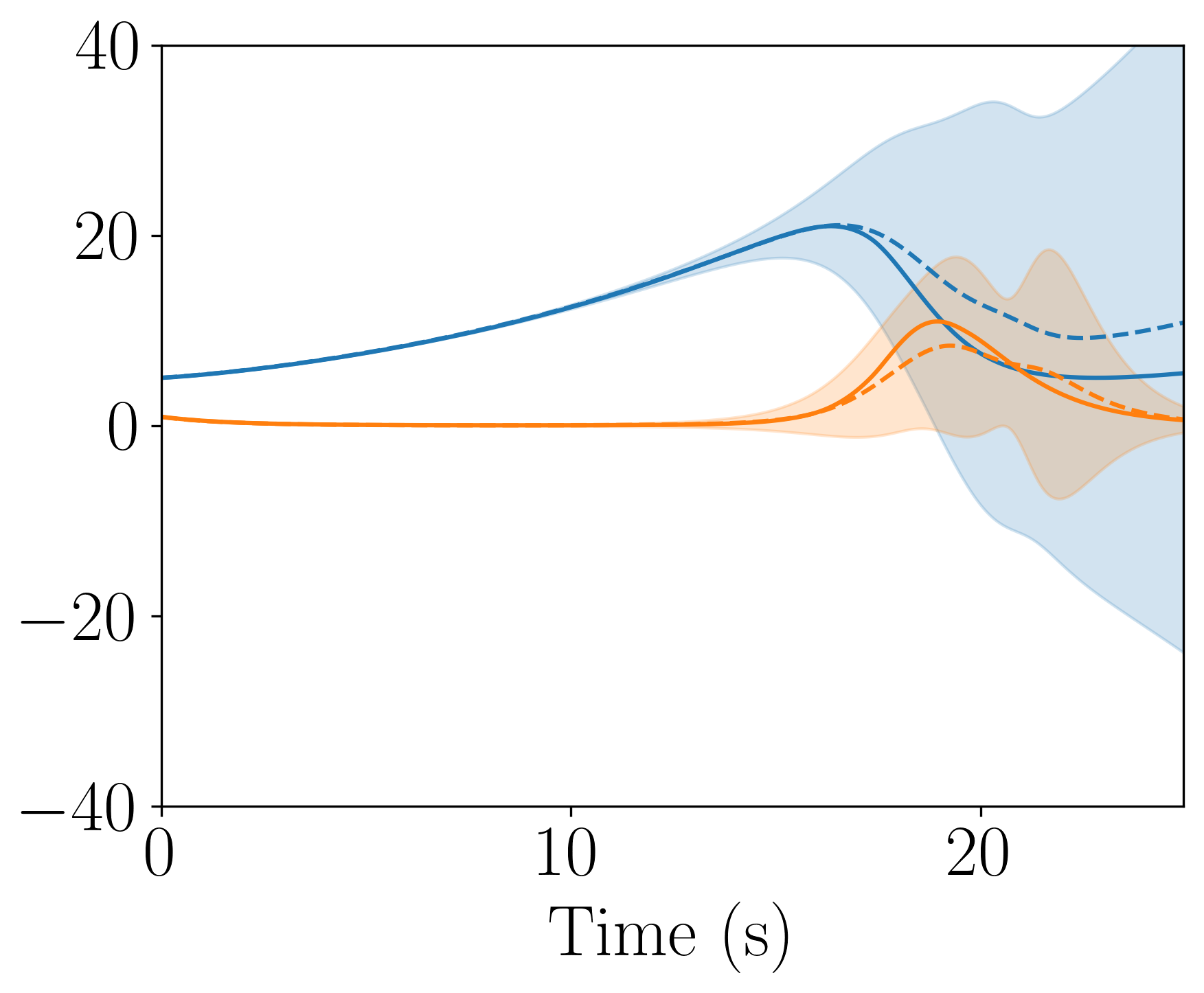}
        \caption{Injection $xy$ in the first layer}
    \end{subfigure}
    \begin{subfigure}[b]{0.475\linewidth}
        \includegraphics[width=\linewidth]{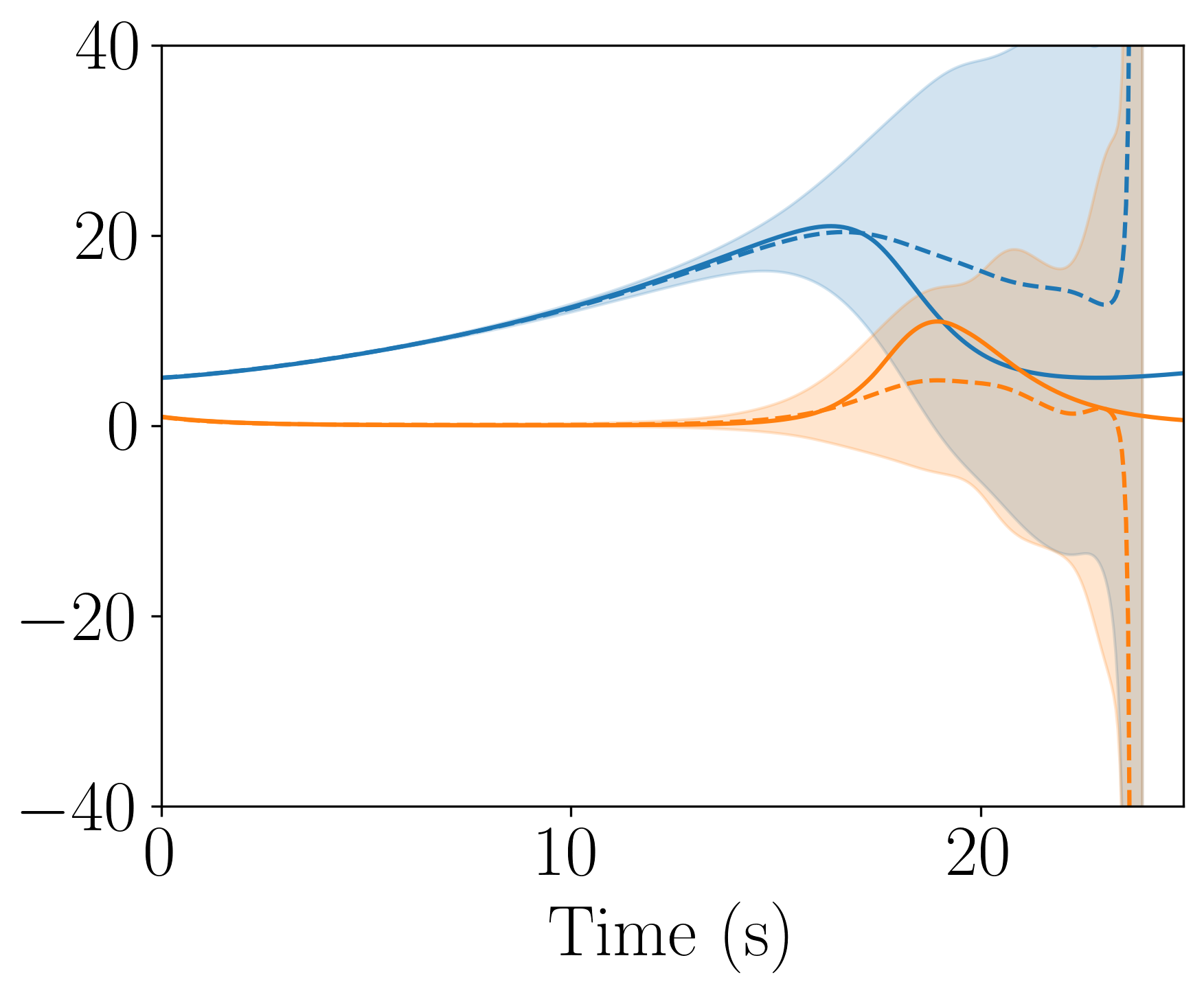}
        \caption{Injection $xy$ in second layer}
    \end{subfigure}
    \begin{subfigure}[b]{0.475\linewidth}
        \includegraphics[width=\linewidth]{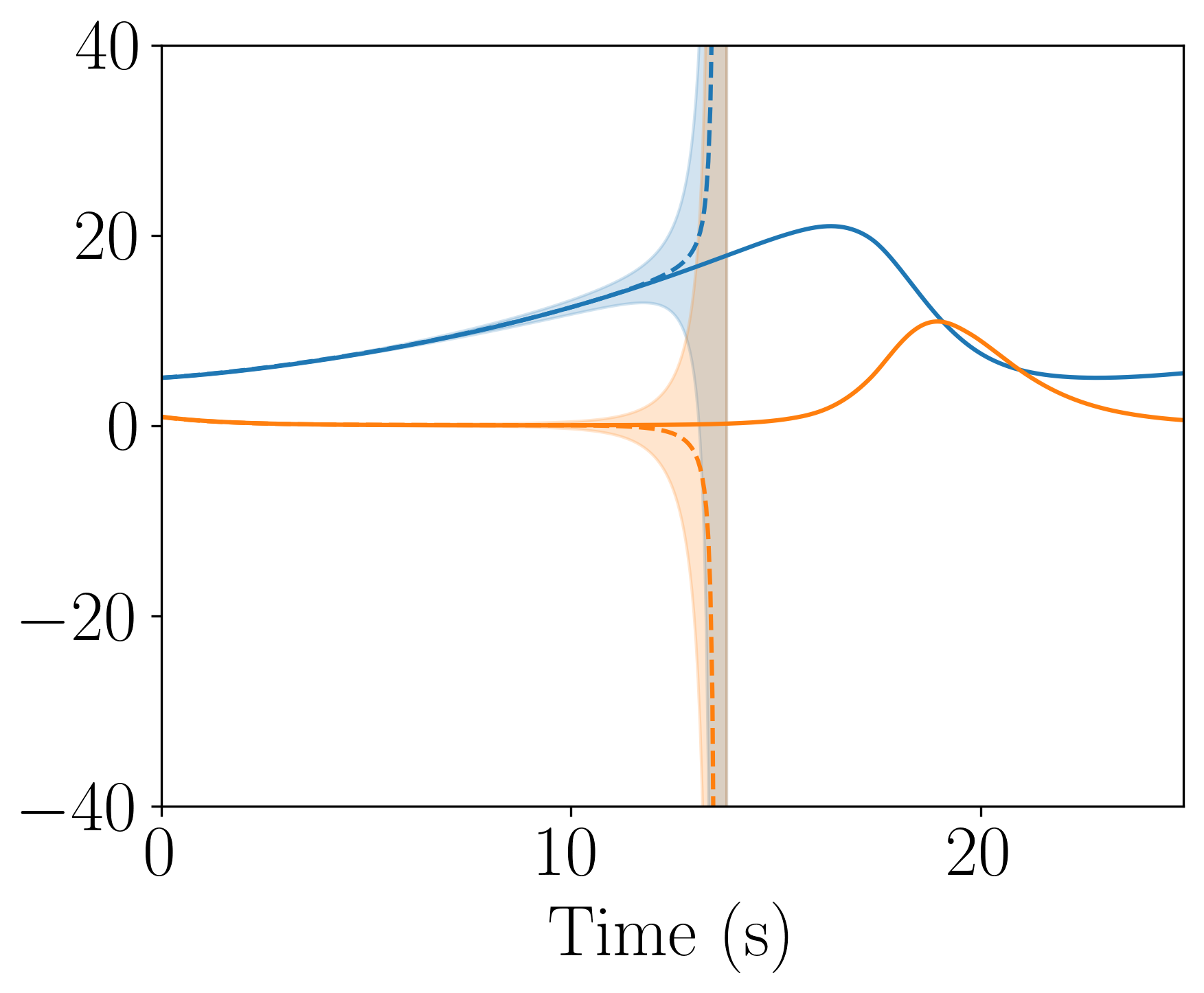}
        \caption{Injection $xy$ in third layer}
    \end{subfigure}
    \caption{Rolling forecast for the Lotka-Volterra system with and without injection at different layers. The best results are achieved through knowledge injection in the first layer. Injecting into the second and third layers appears to cause blowup.}
    \label{fig:lotkavolterra_reg_forecast}
\end{figure}

Figure \ref{fig:lotkavolterra_training_loss_comparison} shows that injecting the $xy$ term in any layer caused the networks to reach a lower validation loss more quickly. This improvement was greatest when the injection was placed in the first hidden layer. Figure \ref{fig:lotkavolterra_reg_forecast} shows the mean rolling forecast of the ensembles with a 95\% confidence interval. Injection in the first layer significantly improves the accuracy and the uncertainty of the forecast. However, the forecast quickly blows up for the models with second and third layer injections. This is especially pronounced for the third layer injection.

\subsection{Duffing system with \texorpdfstring{$x^3$}{x3} injection}
\label{sec:duffing}
\begin{figure}
    \centering
    \begin{adjustbox}{max width=0.75\linewidth}
        \begin{tikzpicture}
            \begin{customlegend}[legend columns=4,legend style={draw=none, column sep=1ex},legend entries={No injection, $\cos(\omega t)$ layer 1, $\cos(\omega t)$ layer 2, $\cos(\omega t)$ layer 3, $xy$ layer 1, $xy$ layer 2, $xy$ layer 3}]
                \addlegendimage{black,sharp plot}
                \addlegendimage{Tblue,sharp plot}
                \addlegendimage{Torange,sharp plot}
                \addlegendimage{Tgreen,sharp plot}
                \addlegendimage{Tred,sharp plot}
                \addlegendimage{Tpurple,sharp plot}
                \addlegendimage{Tbrown,sharp plot}
            \end{customlegend}
        \end{tikzpicture}
    \end{adjustbox}

    \begin{subfigure}[b]{0.475\linewidth}
        \includegraphics[width=\linewidth]{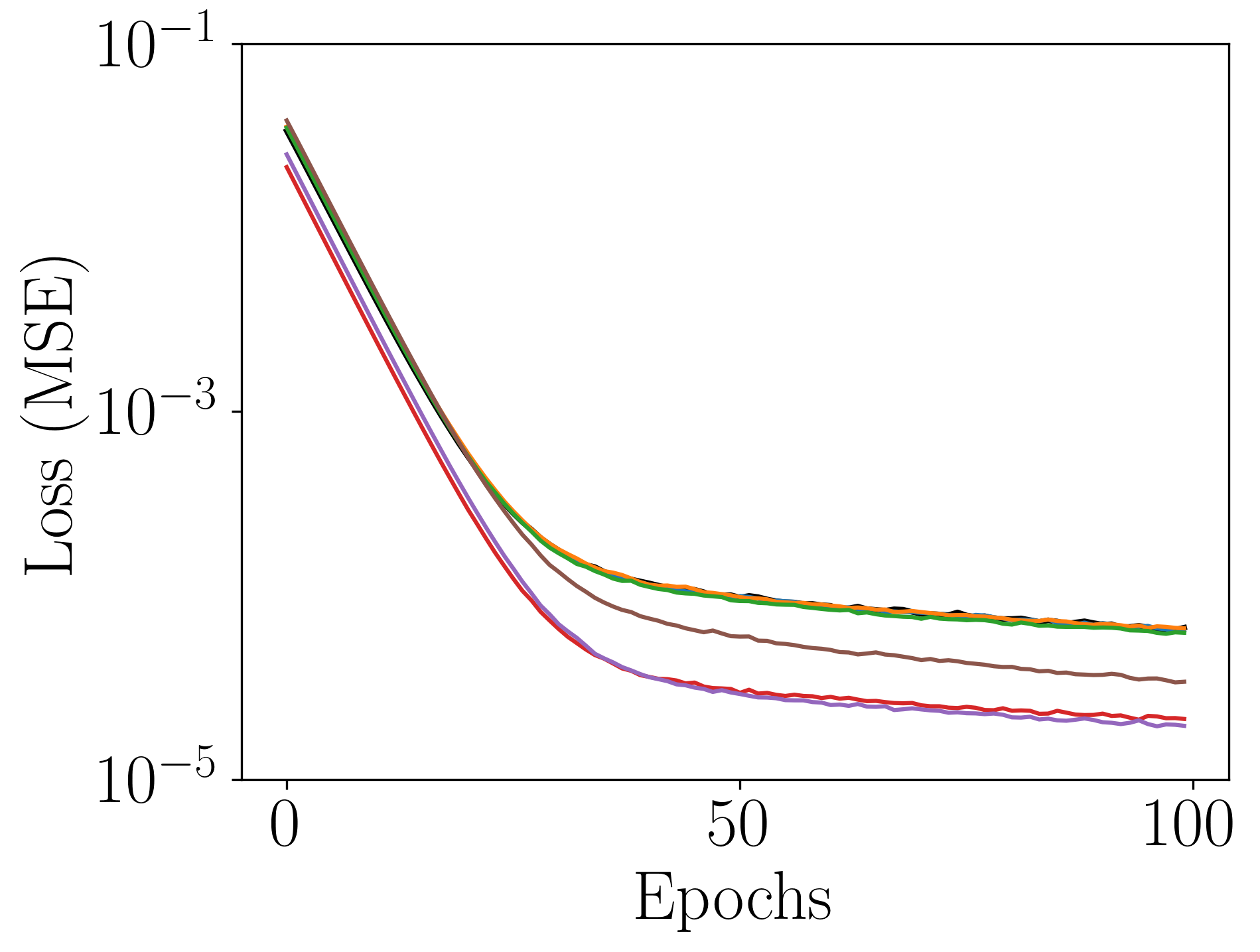}
        \caption{Training loss}
    \end{subfigure}
    \begin{subfigure}[b]{0.475\linewidth}
        \includegraphics[width=\linewidth]{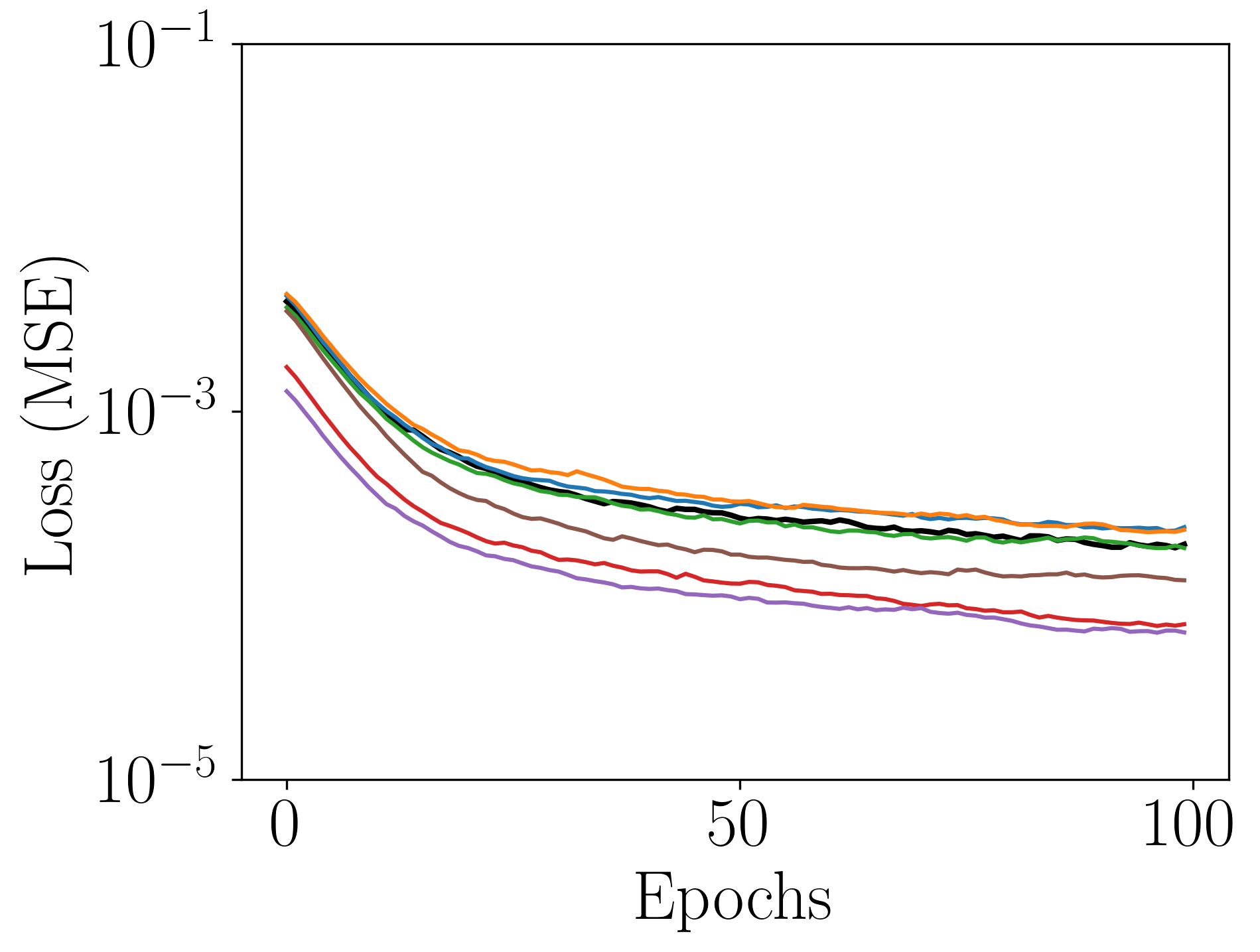}
        \caption{Validation loss}
    \end{subfigure}
    \caption{Comparison of the training loss for the Duffing oscillator for different injection configurations.}
    \label{fig:duffing_training_loss_comparison}
\end{figure}

Figure \ref{fig:duffing_training_loss_comparison} compares the training and validation loss of the models injected with $x^3$ and $\cos(\omega t)$ terms. Both training and validation loss are significantly improved with the $x^3$ injection, while $\cos(\omega t)$ appears to have little effect.

\begin{figure}
    \centering
    \begin{adjustbox}{max width=0.5\linewidth}
        \begin{tikzpicture}
            \begin{customlegend}[legend columns=3,legend style={draw=none, column sep=1ex},legend entries={Truth,Prediction,95\% conf.,$x$,$y$}]
                \addlegendimage{black, sharp plot}
                \addlegendimage{black, dashed,sharp plot}
                \addlegendimage{black!20, fill=black!20, area legend}
                \addlegendimage{Tblue,sharp plot}
                \addlegendimage{Torange,sharp plot}
            \end{customlegend}
        \end{tikzpicture}
    \end{adjustbox}

    \begin{subfigure}[b]{0.475\linewidth}
        {\includegraphics[width=\linewidth]{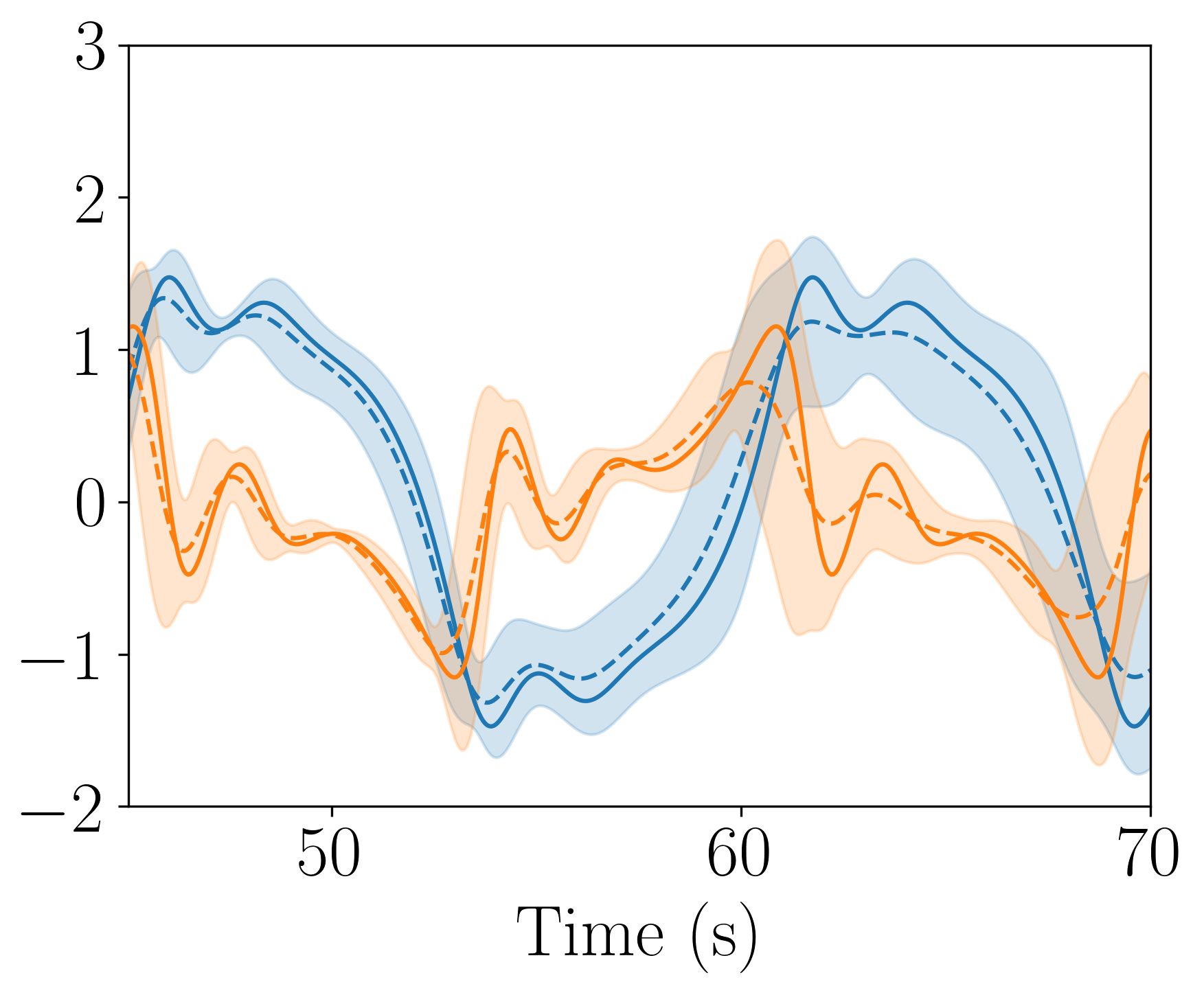}}
        \caption{No injection}
    \end{subfigure}
    \begin{subfigure}[b]{0.475\linewidth}
        \includegraphics[width=\linewidth]{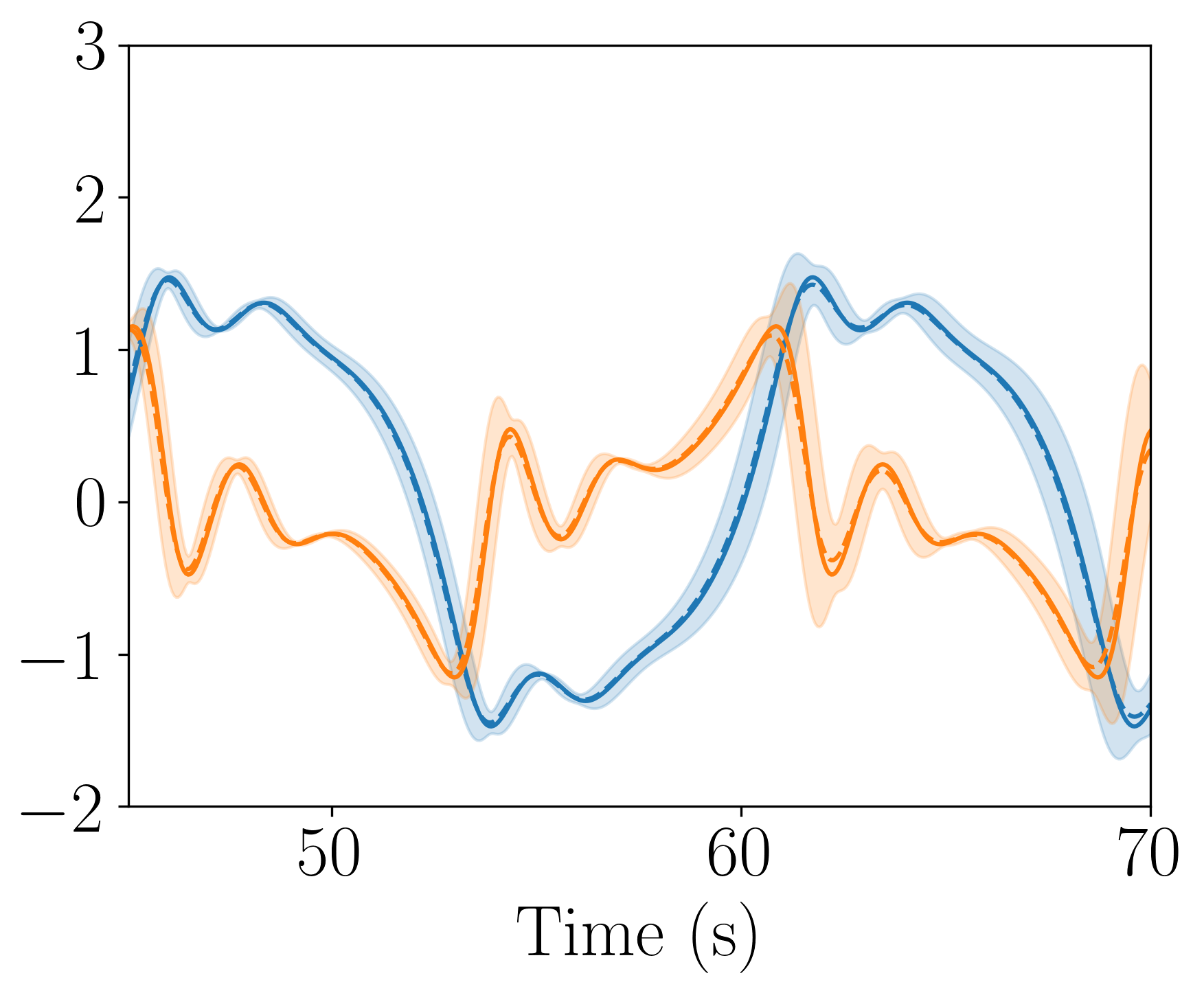}
        \caption{Injection $x^3$ in the first layer}
    \end{subfigure}
    \begin{subfigure}[b]{0.475\linewidth}
        \includegraphics[width=\linewidth]{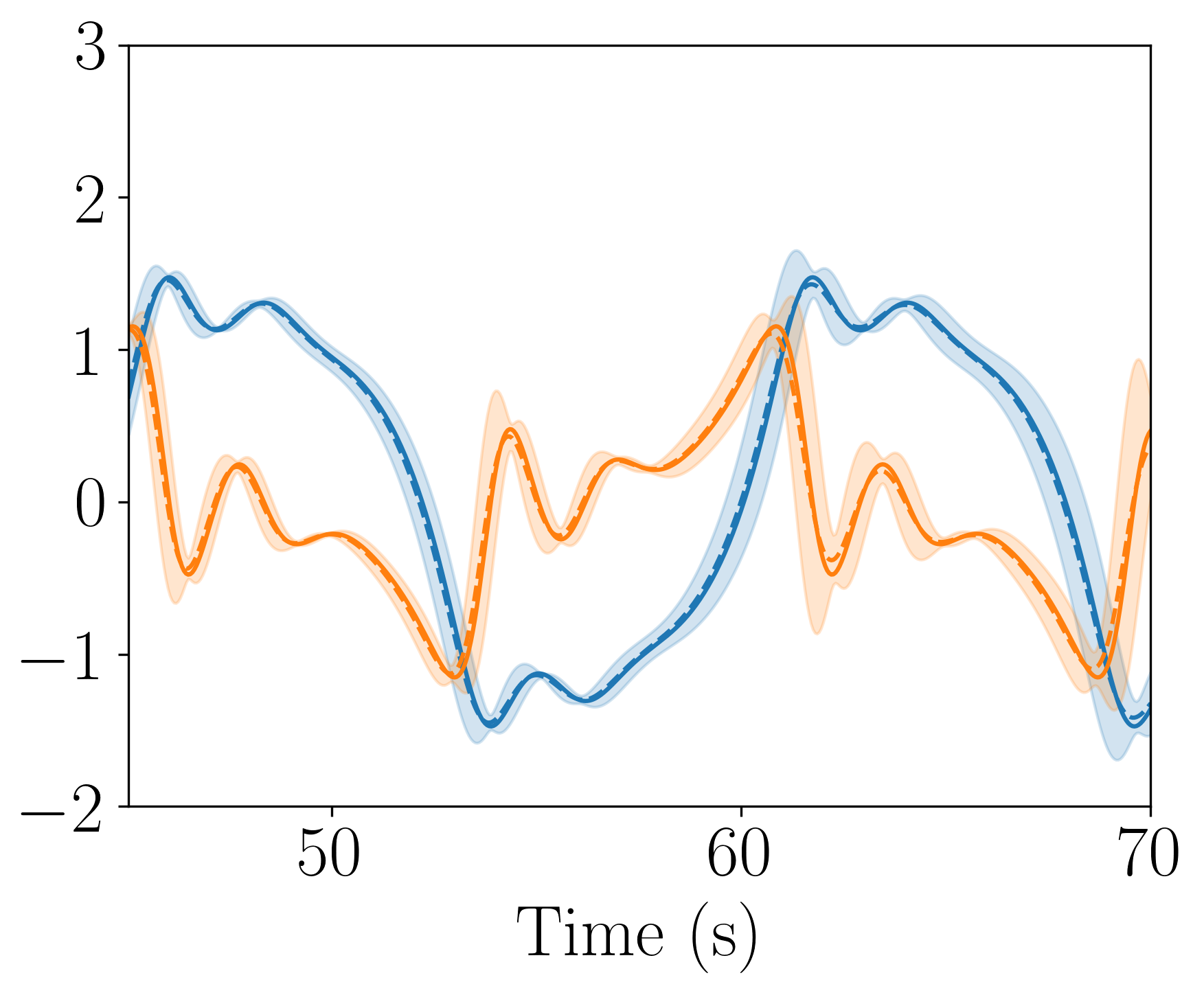}
        \caption{Injection $x^3$ in second layer}
    \end{subfigure}
    \begin{subfigure}[b]{0.475\linewidth}
        {\includegraphics[width=\linewidth]{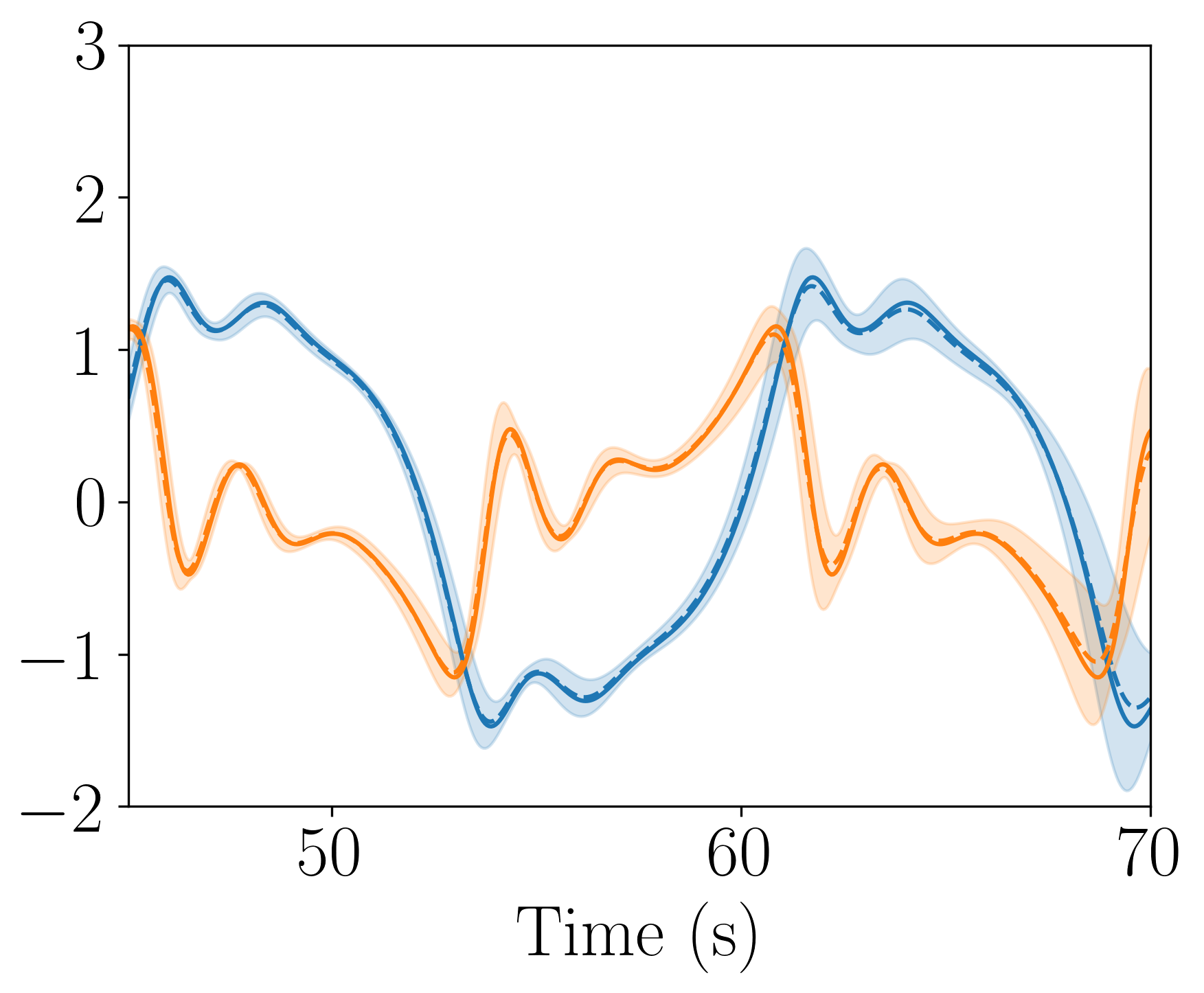}}
        \caption{Injection $x^3$ in third layer}
    \end{subfigure}
    \caption{Rolling forecast for the Duffing oscillator with and without injection at different layers. A small time segment from 45s--70s from the forecast is shown here to highlight the differences between the models.}
    \label{fig:duffing_reg_forecast}
\end{figure}

The predicted trajectories for $x^3$ models can be seen in Figure \ref{fig:duffing_reg_forecast}, while the $\cos(\omega t)$ models have been omitted for brevity. Note that the figure shows a time segment from 75s--100s in order to highlight the differences between the models. We observe that knowledge injection improves the accuracy and model uncertainty in all cases, and all ensembles perform similarly.

It is interesting that although $\cos(\omega t)$ is available as an input to the network (due to the parameterisation described in Section \ref{sec:duffing-bg}), injecting the same term changes the RFMSE significantly, despite being redundant information. However, this is not reflected in the training and validation loss, where the baseline model and the models injected with $\cos({\omega t})$ appear to have nearly identical training characteristics. We found that when forecasting over longer periods, the $\cos({\omega t})$ models yielded unstable predictions with a higher rate of blowup, while the other models tended to decay instead.

\subsection{Van der Pol system with \texorpdfstring{$x^2y$}{x2y} injection}

\begin{figure}
    \centering
    \begin{adjustbox}{max width=0.75\linewidth}
        \begin{tikzpicture}
            \begin{customlegend}[legend columns=4,legend style={draw=none, column sep=1ex},legend entries={No injection, $x^2y$ layer 1, $x^2y$ layer 2, $x^2y$ layer 3, $xy$ layer 1, $xy$ layer 2, $xy$ layer 3}]
                \addlegendimage{black,sharp plot}
                \addlegendimage{Tblue,sharp plot}
                \addlegendimage{Torange,sharp plot}
                \addlegendimage{Tgreen,sharp plot}
                \addlegendimage{Tred,sharp plot}
                \addlegendimage{Tpurple,sharp plot}
                \addlegendimage{Tbrown,sharp plot}
            \end{customlegend}
        \end{tikzpicture}
    \end{adjustbox}

    \begin{subfigure}[b]{0.475\linewidth}
        \includegraphics[width=\linewidth]{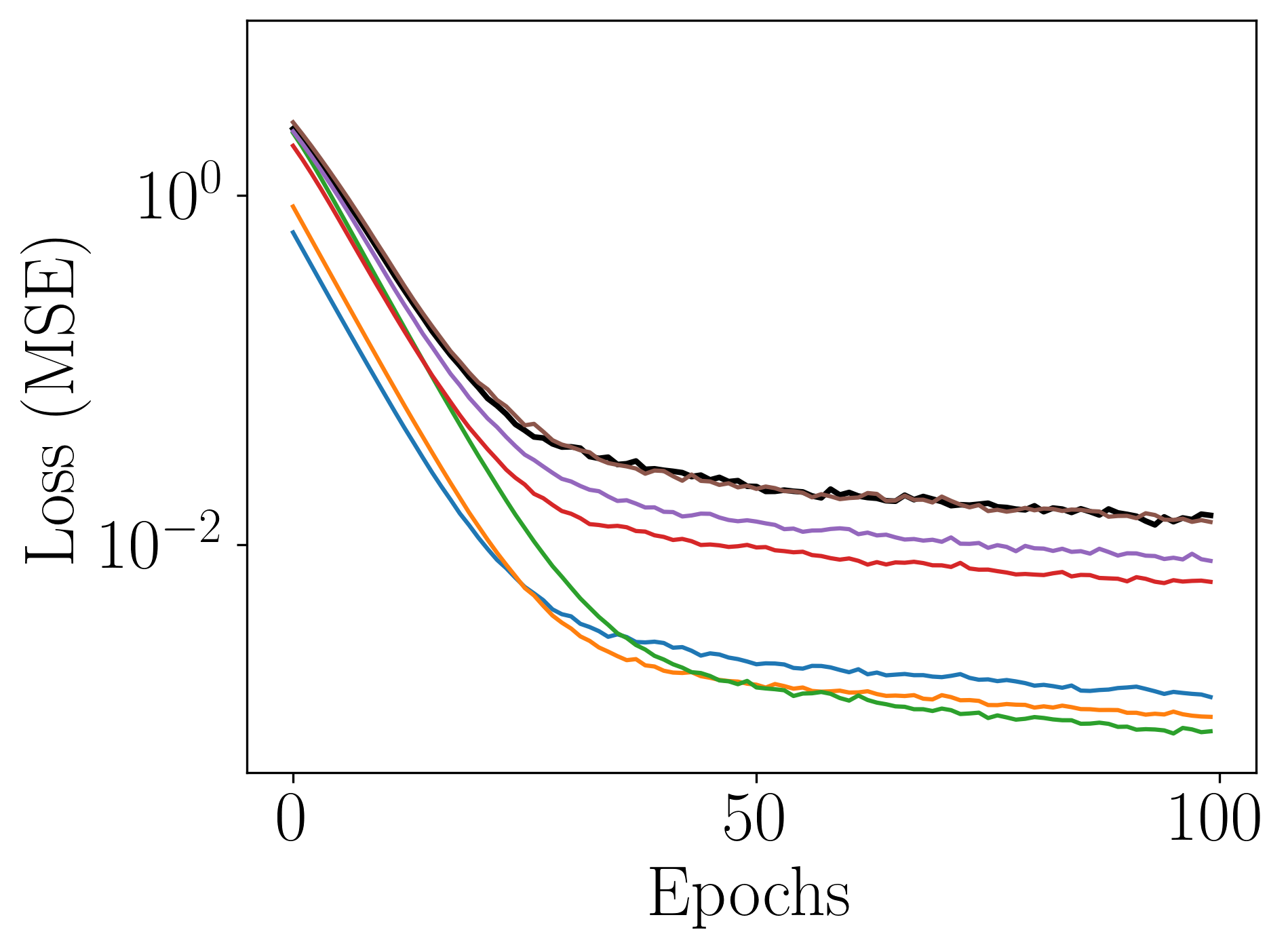}
        \caption{Training loss}
    \end{subfigure}
    \begin{subfigure}[b]{0.475\linewidth}
        \includegraphics[width=\linewidth]{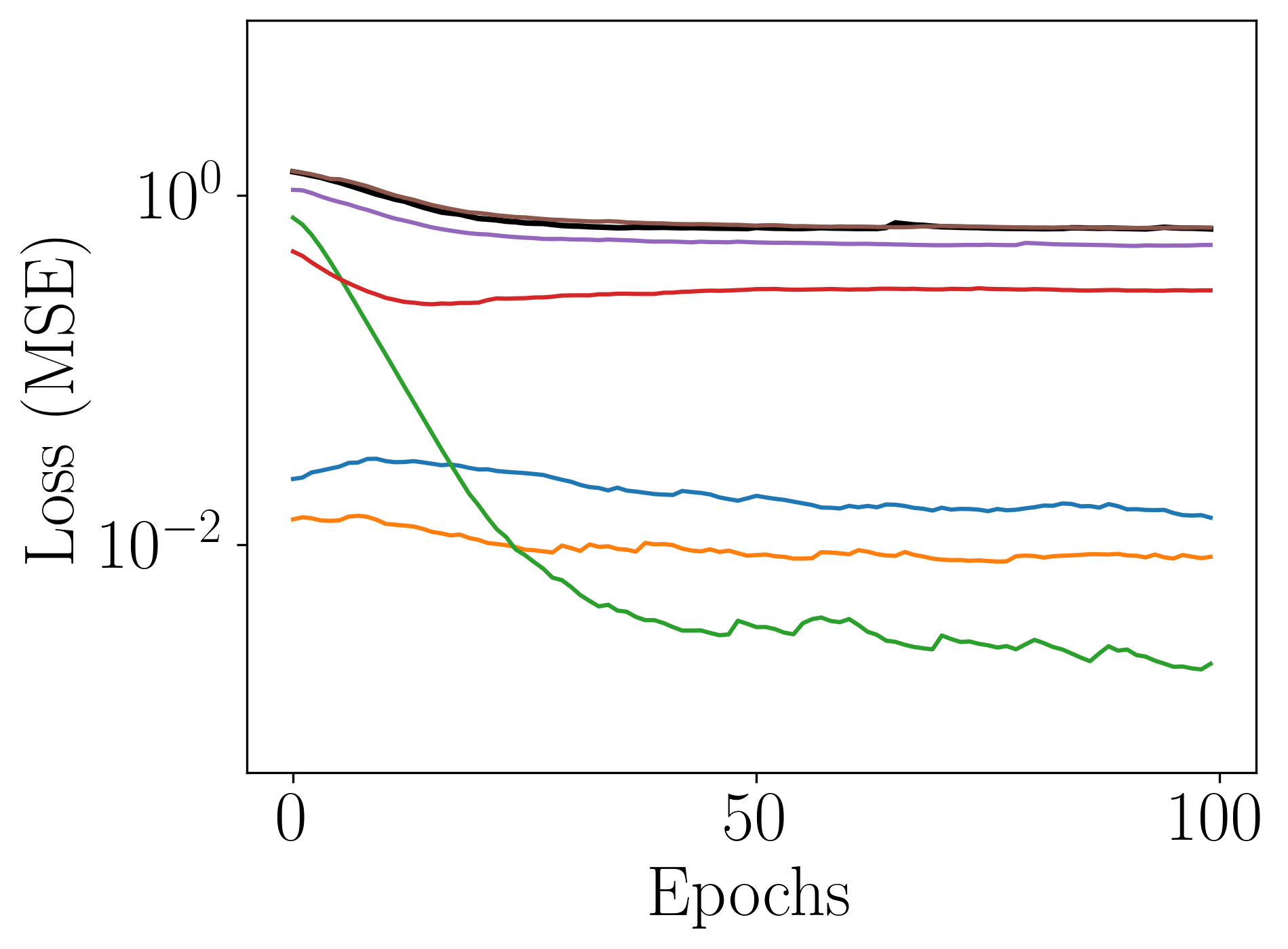}
        \caption{Validation loss}
    \end{subfigure}
    \caption{Comparison of the training and validation loss for the Van der Pol oscillator for different injection configurations.}
    \label{fig:vanderpol_training_loss_comparison}
\end{figure}

For this system, the functions $x^2y$ and $x^2$ were injected in all three layers. Figure \ref{fig:vanderpol_training_loss_comparison} shows how the $x^2y$ injected models improve validation loss by 1-2 orders of magnitude better than the model without injection.

\begin{figure}[t]
    \centering
    \begin{adjustbox}{max width=0.5\linewidth}
        \begin{tikzpicture}
            \begin{customlegend}[legend columns=3,legend style={draw=none, column sep=1ex},legend entries={Truth,Prediction,95\% conf.,$x$,$y$}]
                \addlegendimage{black, sharp plot}
                \addlegendimage{black, dashed,sharp plot}
                \addlegendimage{black!20, fill=black!20, area legend}
                \addlegendimage{Tblue,sharp plot}
                \addlegendimage{Torange,sharp plot}
            \end{customlegend}
        \end{tikzpicture}
    \end{adjustbox}

    \begin{subfigure}[b]{0.475\linewidth}
        {\includegraphics[width=\linewidth]{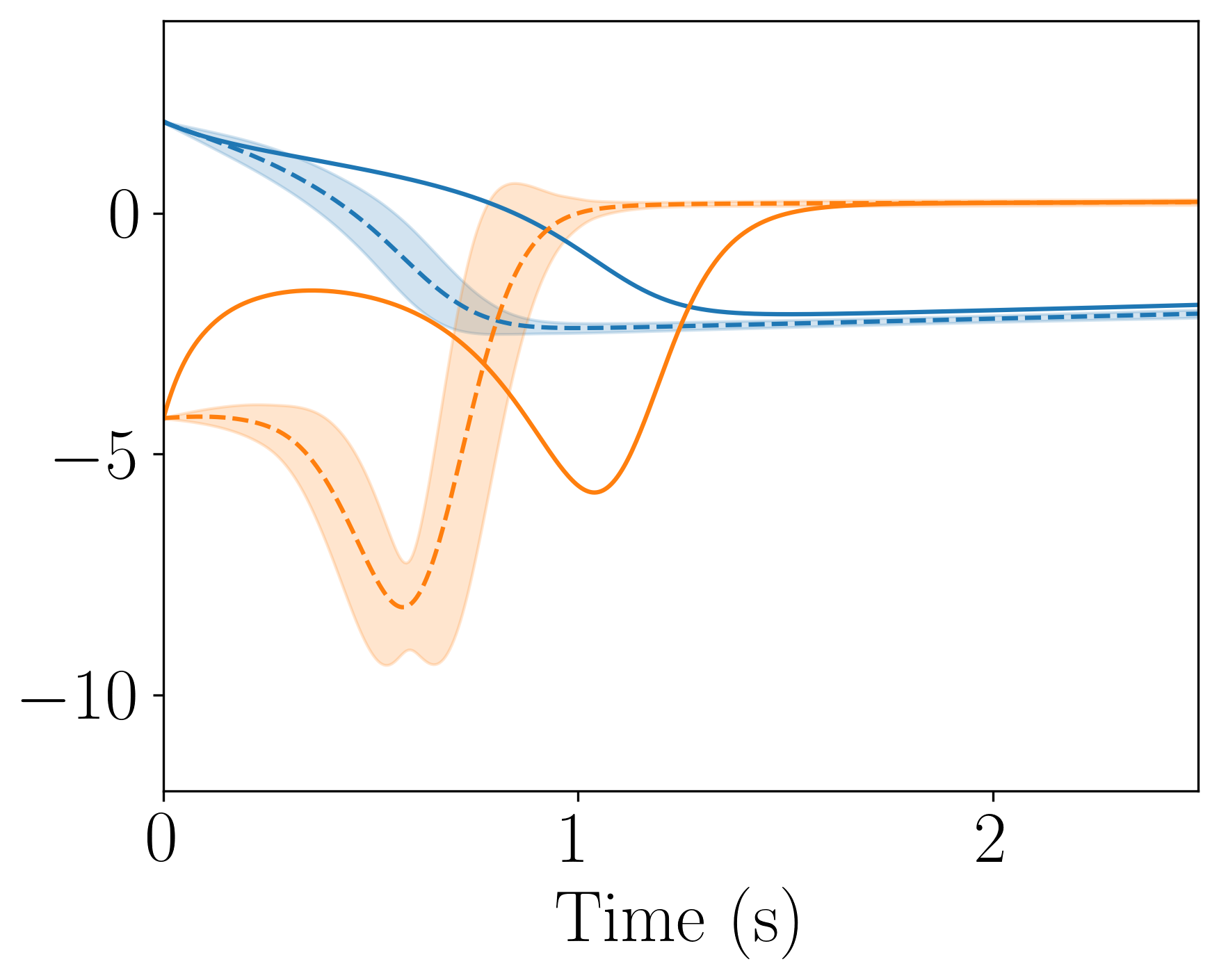}}
        \caption{No injection}
    \end{subfigure}
    \begin{subfigure}[b]{0.475\linewidth}
        \includegraphics[width=\linewidth]{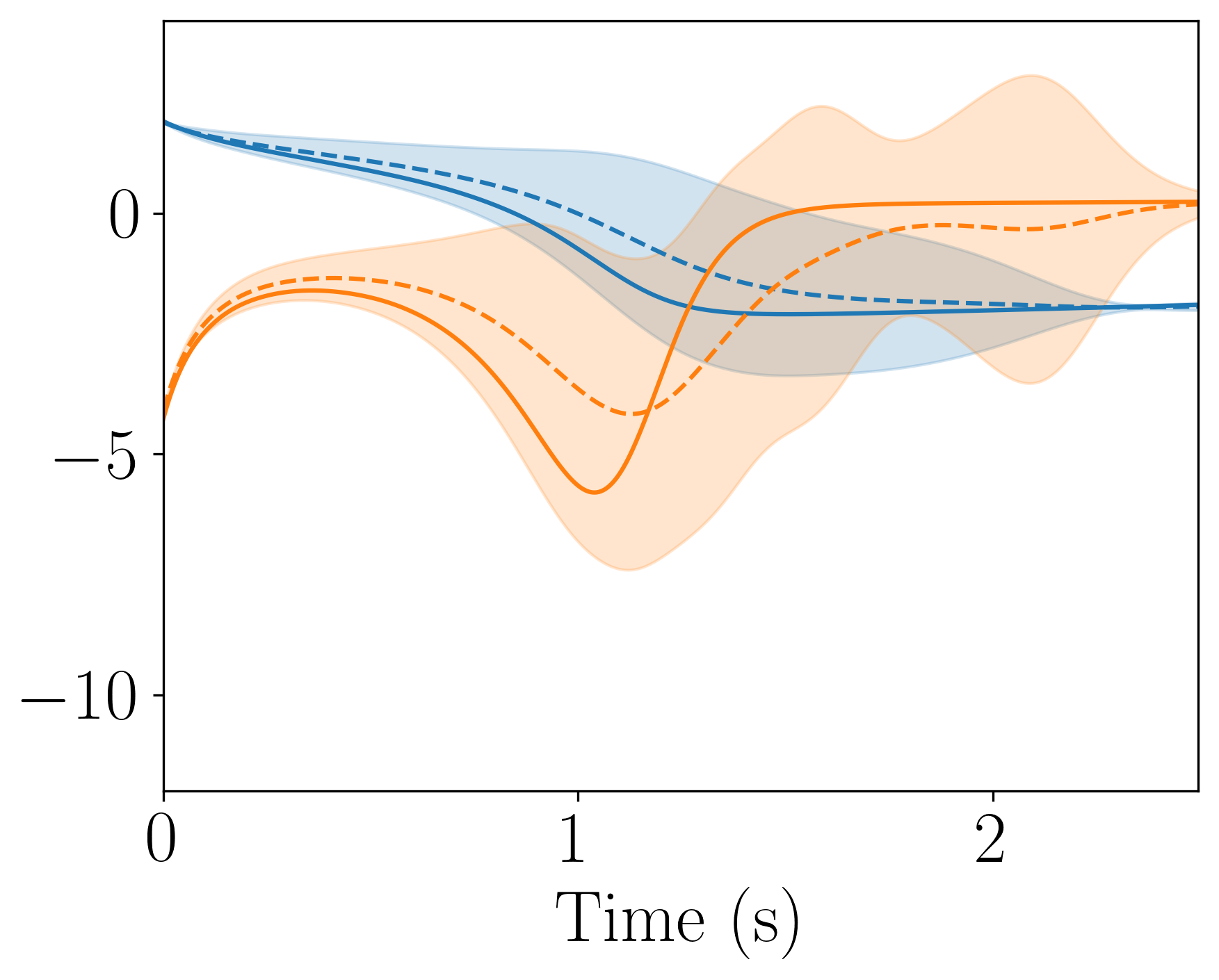}
        \caption{Injection $x^2y$ in the first layer}
    \end{subfigure}
    \begin{subfigure}[b]{0.475\linewidth}
        \includegraphics[width=\linewidth]{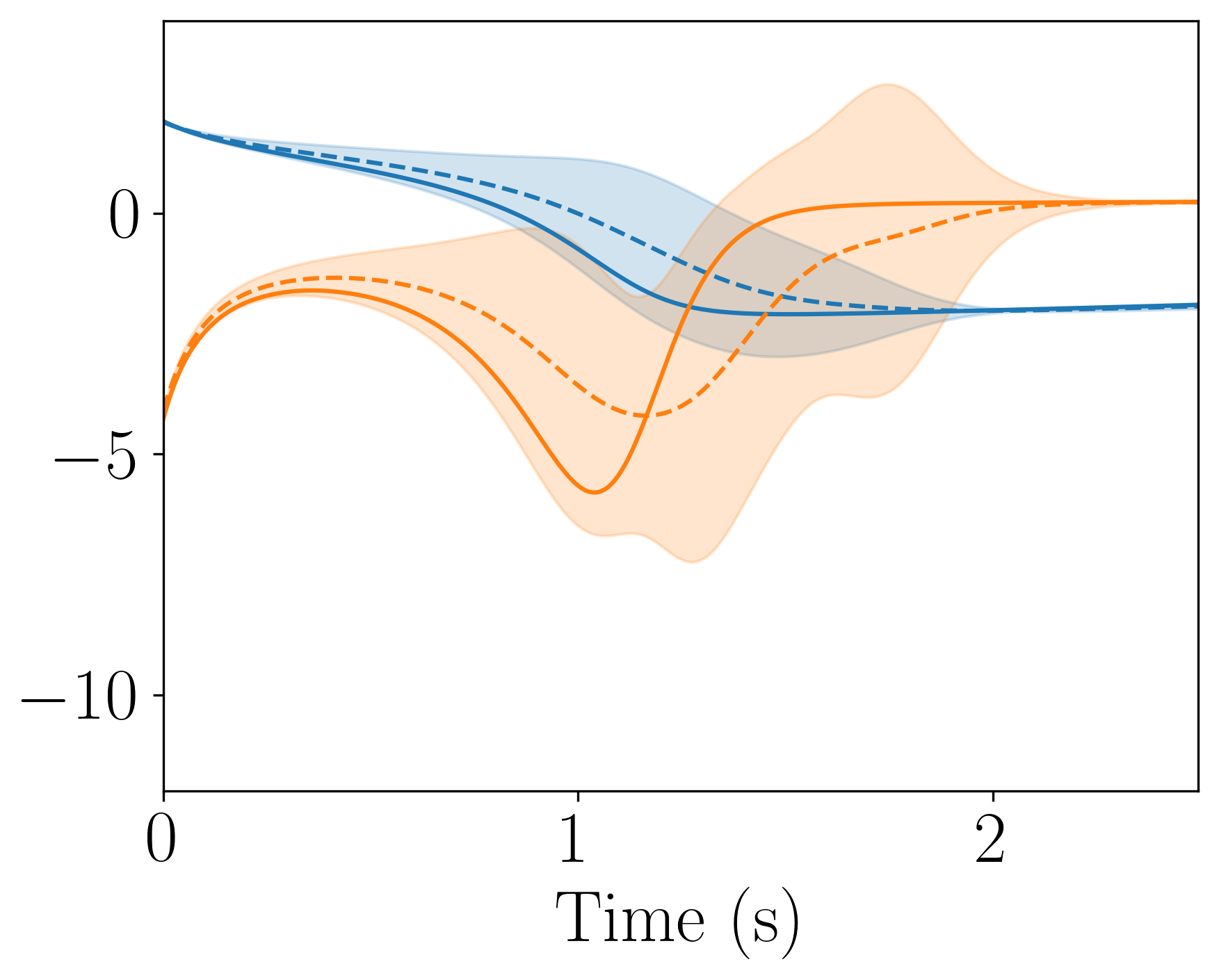}
        \caption{Injection $x^2y$ in second layer}
    \end{subfigure}
    \begin{subfigure}[b]{0.475\linewidth}
        \includegraphics[width=\linewidth]{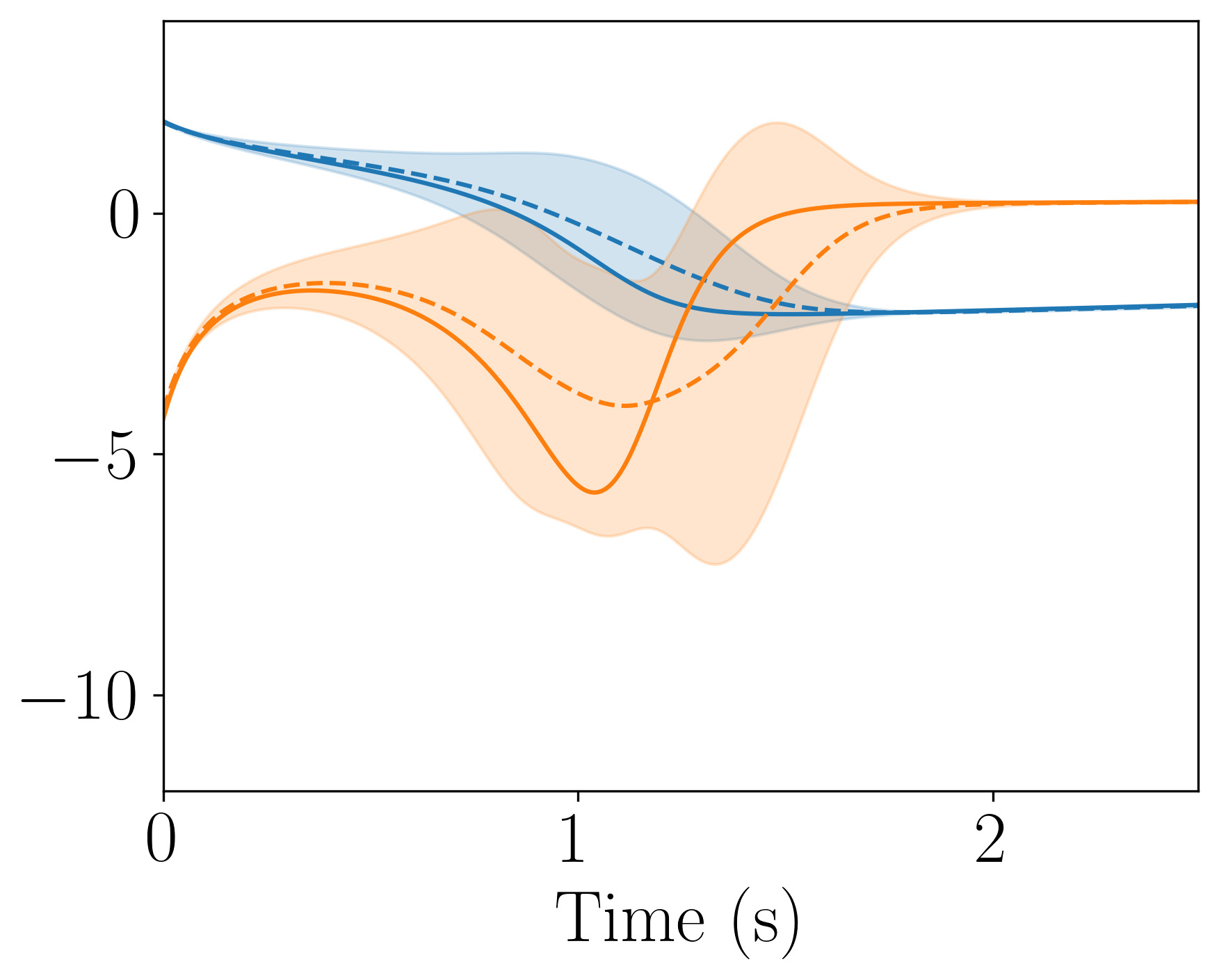}
        \caption{Injection $x^2y$ in third layer}
    \end{subfigure}
    \caption{Rolling forecast for the Van der Pol oscillator with and without injection at different layers. }
    \label{fig:vanderpol_reg_forecast}
\end{figure}

This large improvement in loss can be understood by inspecting Figure \ref{fig:vanderpol_reg_forecast}, which shows a rolling forecast on the test trajectory near a fast transient. The figure shows that the models without injection fail to properly capture the transient, and also fail to converge to the slow dynamics afterwards. The long duration of the slow dynamics likely leads to a large build-up of error over the course of the full trajectory.

Surprisingly, the ensemble with no knowledge injection also exhibits very low model uncertainty. This might again be explained by the fast and slow dynamics of the Van der Pol oscillator. The dataset is likely unbalanced, dominated by slower varying states due to the longer duration of the slow dynamics. This could cause poor performance on the fast transients, which can be seen as relatively rare events.

Figure \ref{fig:vanderpol_reg_forecast} shows that this is greatly improved through knowledge injection. All injected models track the transient more closely, reach the correct value for the slow dynamics, and the true trajectory is within the 95\% confidence intervals for all model classes. The model uncertainty is naturally increased at the transient, and appears to shrink to zero when the slow dynamics set in.

\subsection{Lorenz System with \texorpdfstring{$xy$}{xy} injection}
\begin{figure}[H]
    \centering
    \begin{adjustbox}{max width=0.75\linewidth}
        \begin{tikzpicture}
            \begin{customlegend}[legend columns=4,legend style={draw=none, column sep=1ex},legend entries={No injection, $xy$ layer 1, $xy$ layer 2, $xy$ layer 3}]
                \addlegendimage{black,sharp plot}
                \addlegendimage{Tblue,sharp plot}
                \addlegendimage{Torange,sharp plot}
                \addlegendimage{Tgreen,sharp plot}
            \end{customlegend}
        \end{tikzpicture}
    \end{adjustbox}

    \begin{subfigure}[b]{0.475\linewidth}
        \includegraphics[width=\linewidth]{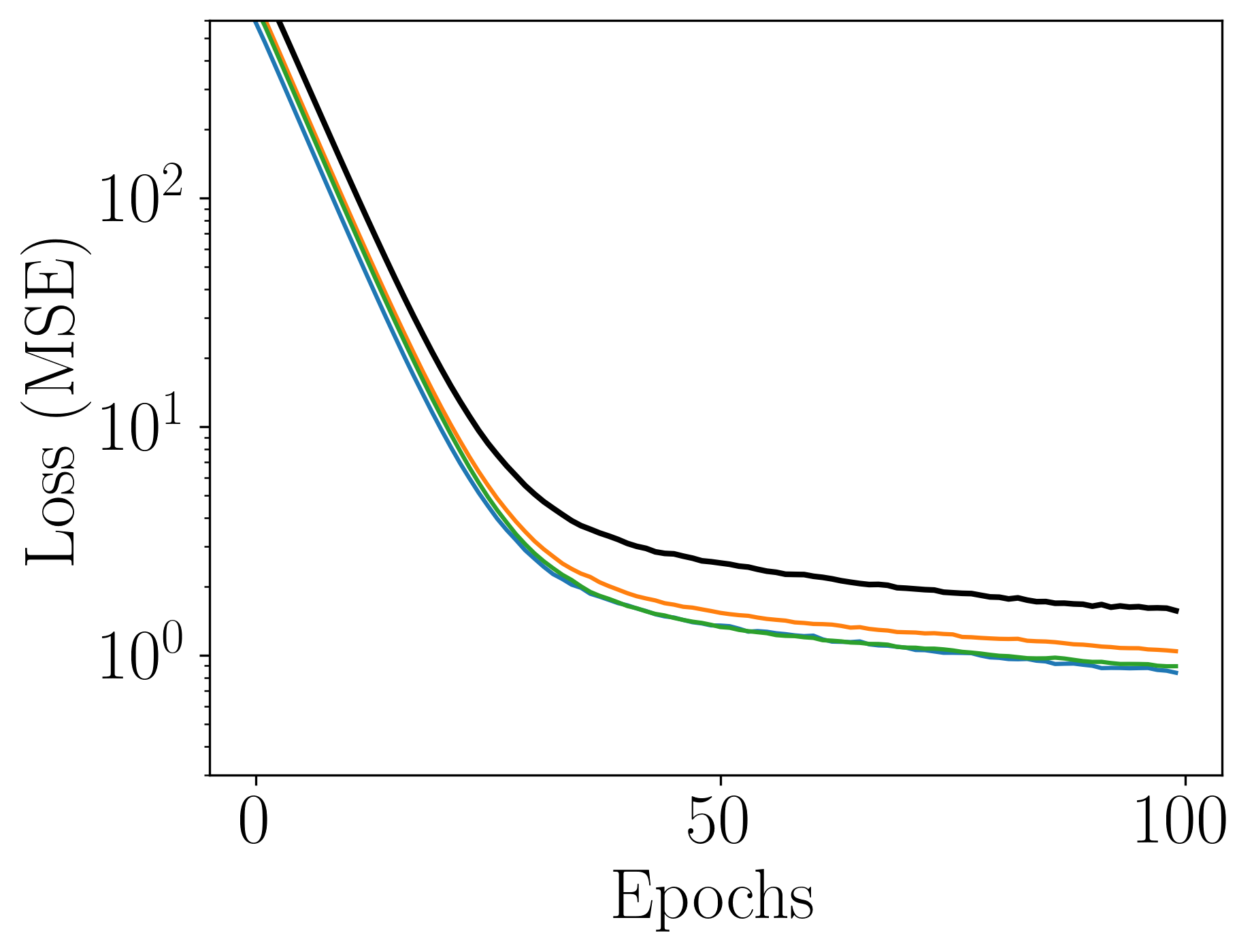}
        \caption{Training loss}
    \end{subfigure}
    \begin{subfigure}[b]{0.475\linewidth}
        \includegraphics[width=\linewidth]{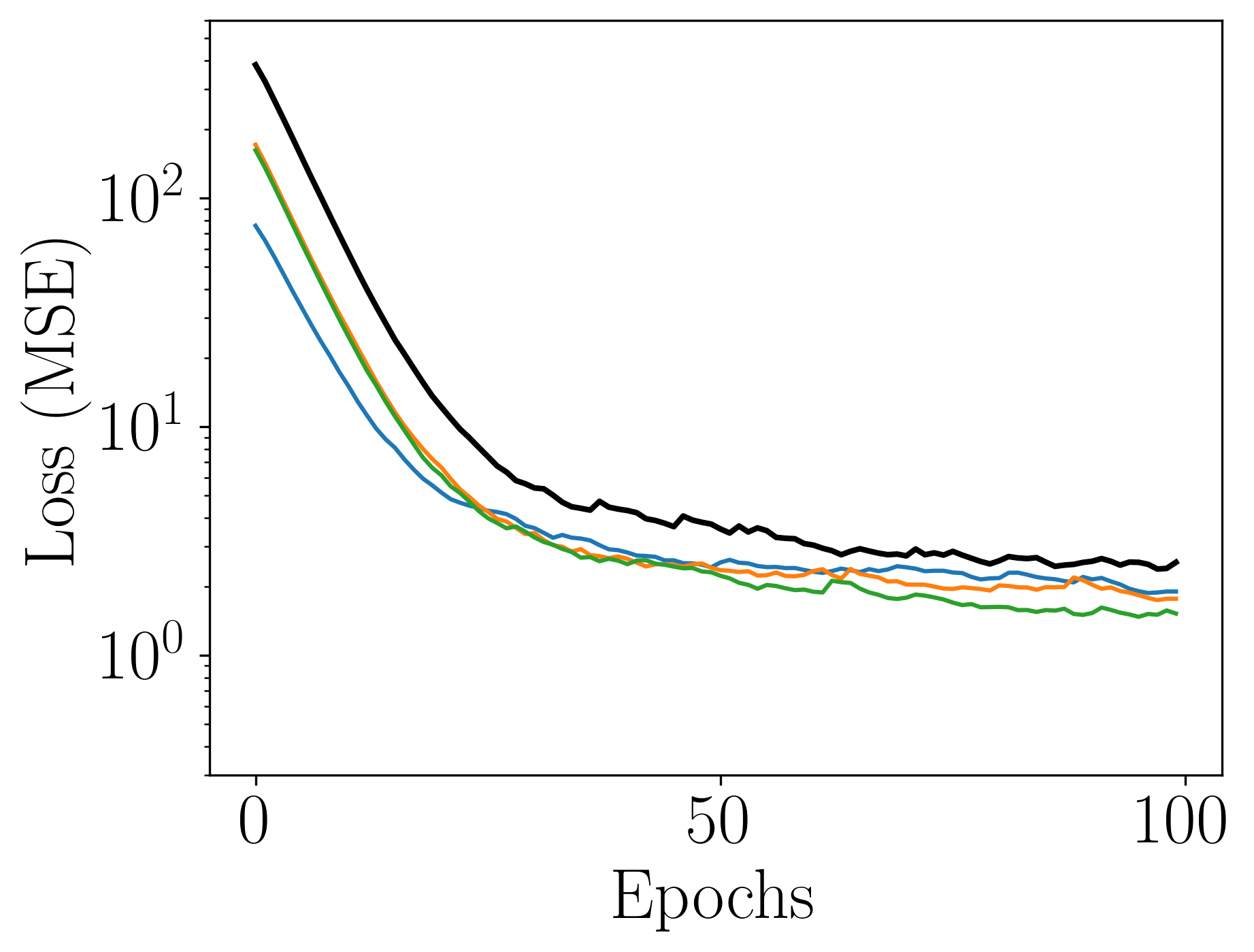}
        \caption{Validation loss}
    \end{subfigure}
    \caption{Comparison of the training and validation loss for the Lorenz system for different injection configurations.}
    \label{fig:lorentz_training_loss_comparison}
\end{figure}

\begin{figure}
    \centering
    \begin{adjustbox}{max width=0.5\linewidth}
        \begin{tikzpicture}
            \begin{customlegend}[legend columns=3,legend style={draw=none, column sep=1ex},legend entries={Truth,Prediction,95\% conf.,$x$,$y$,$z$}]
                \addlegendimage{black, sharp plot}
                \addlegendimage{black, dashed,sharp plot}
                \addlegendimage{black!20, fill=black!20, area legend}
                \addlegendimage{Tblue,sharp plot}
                \addlegendimage{Torange,sharp plot}
                \addlegendimage{Tgreen,sharp plot}
            \end{customlegend}
        \end{tikzpicture}
    \end{adjustbox}

    \begin{subfigure}[b]{0.475\linewidth}
        {\includegraphics[width=\linewidth]{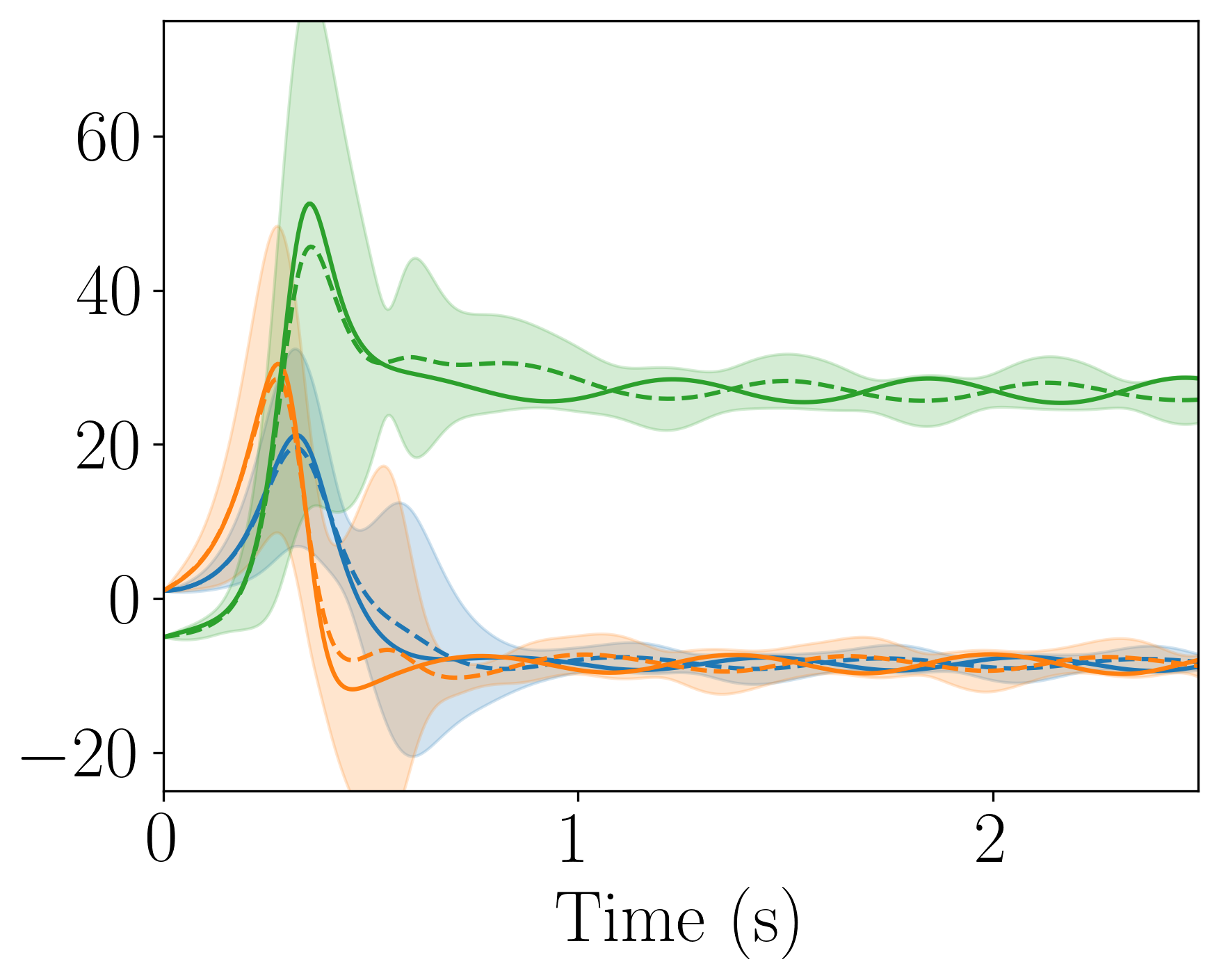}}
        \caption{No injection}
    \end{subfigure}
    \begin{subfigure}[b]{0.475\linewidth}
        \includegraphics[width=\linewidth]{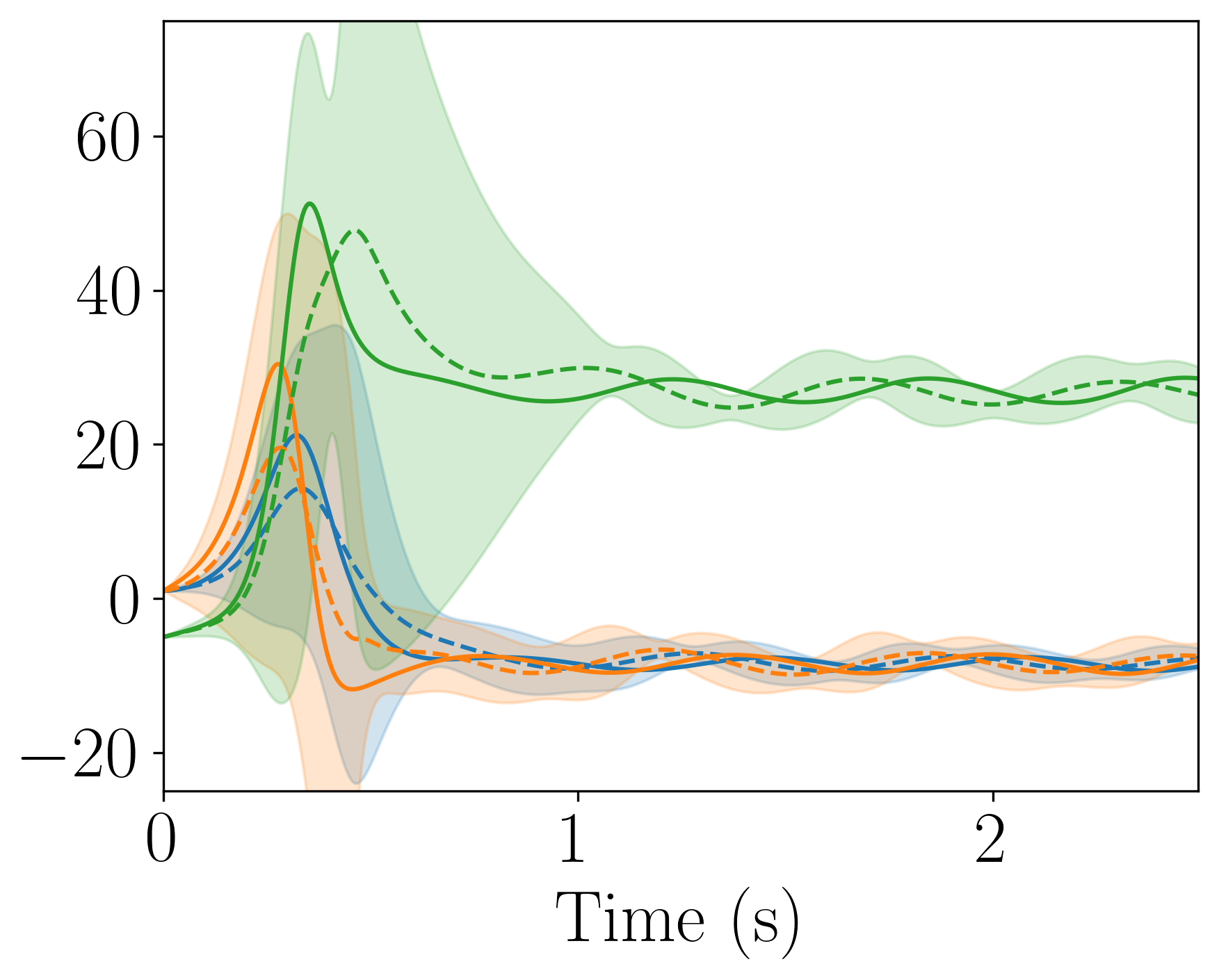}
        \caption{Injection $xy$ in the first layer}
    \end{subfigure}
    \begin{subfigure}[b]{0.475\linewidth}
        \includegraphics[width=\linewidth]{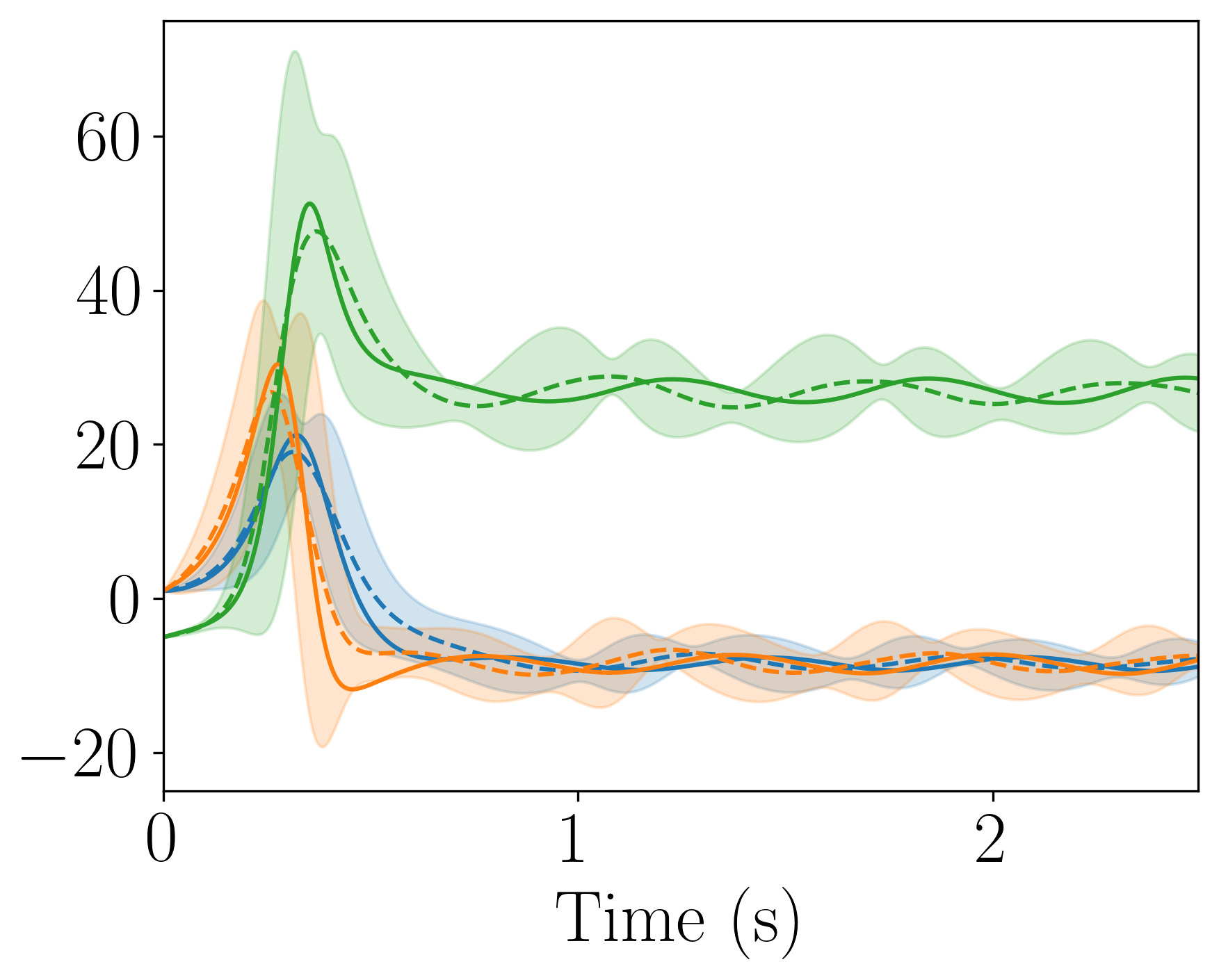}
        \caption{Injection $xy$ in second layer}
    \end{subfigure}
    \begin{subfigure}[b]{0.475\linewidth}
        \includegraphics[width=\linewidth]{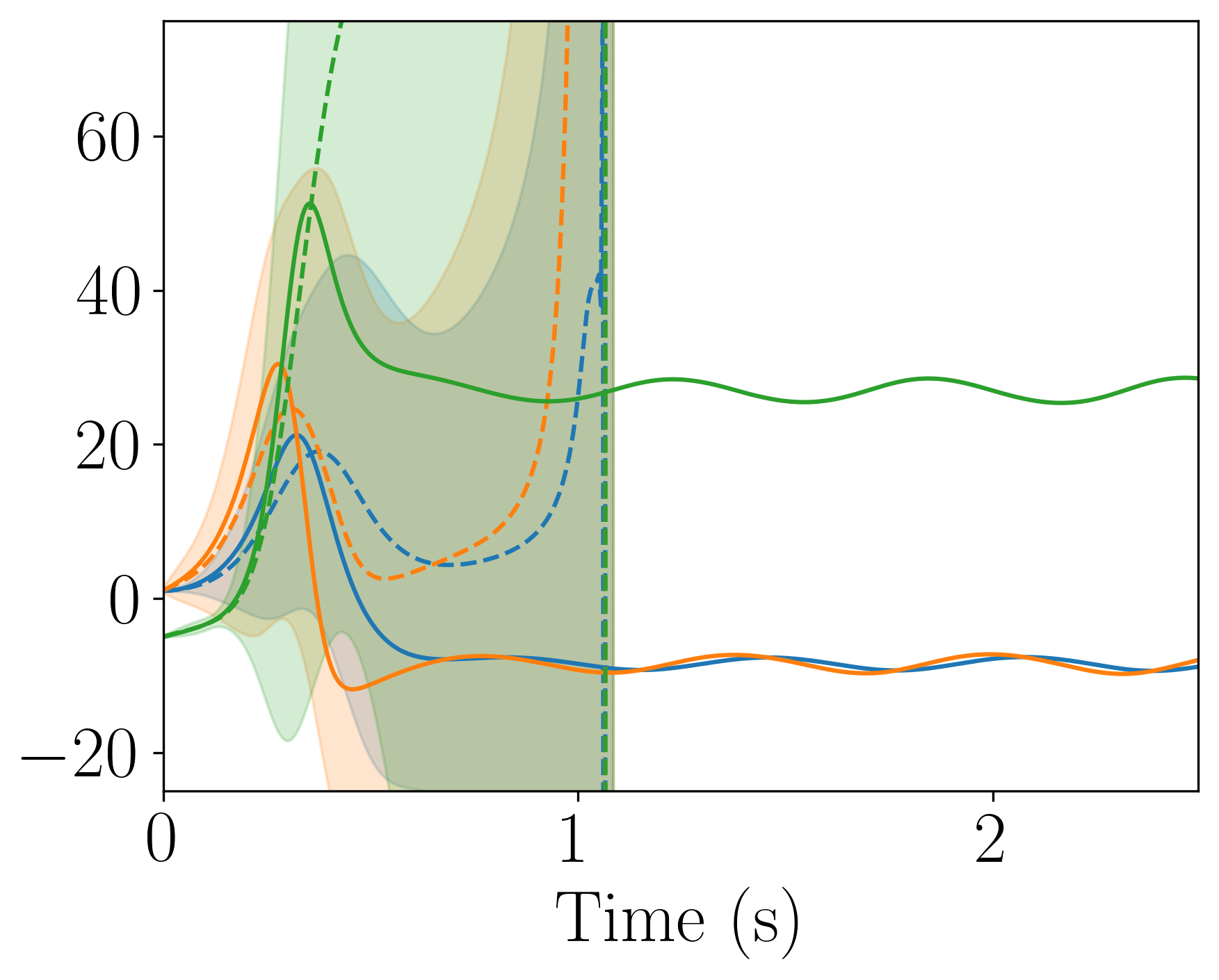}
        \caption{Injection $xy$ in third layer}
    \end{subfigure}
    \caption{Rolling forecast for the Lorenz system without and with injection at different layers. Here the injection is most effective at the second layer, while third layer injection causes the predictions to blow up.}
    \label{fig:lorentz_reg_forecast}
\end{figure}

The function $xy$ was used for knowledge injection. Figure \ref{fig:lorentz_training_loss_comparison} clearly shows that knowledge injection improved the training convergence and final validation loss over 100 epochs. Figure \ref{fig:lorentz_reg_forecast} shows that the ensemble without injection stays close to the true trajectory, but oscillates out of phase with the ground truth. Injection in the first and second layer seems to reduce this lag, and the second layer injection appears to perform marginally better with lower model uncertainty. The predictions from the ensemble with third layer injections quickly diverge.

\subsection{Henon--Heiles system with \texorpdfstring{$xy$}{xy} injection}
\begin{figure}
    \centering
    \begin{adjustbox}{max width=0.75\linewidth}
        \begin{tikzpicture}
            \begin{customlegend}[legend columns=4,legend style={draw=none, column sep=1ex},legend entries={No injection, $xy$ layer 1, $xy$ layer 2, $xy$ layer 3, $y^2$ layer 1, $y^2$ layer 2, $y^2$ layer 3}]
                \addlegendimage{black,sharp plot}
                \addlegendimage{Tblue,sharp plot}
                \addlegendimage{Torange,sharp plot}
                \addlegendimage{Tgreen,sharp plot}
                \addlegendimage{Tred,sharp plot}
                \addlegendimage{Tpurple,sharp plot}
                \addlegendimage{Tbrown,sharp plot}
            \end{customlegend}
        \end{tikzpicture}
    \end{adjustbox}

    \begin{subfigure}[b]{0.475\linewidth}
        \includegraphics[width=\linewidth]{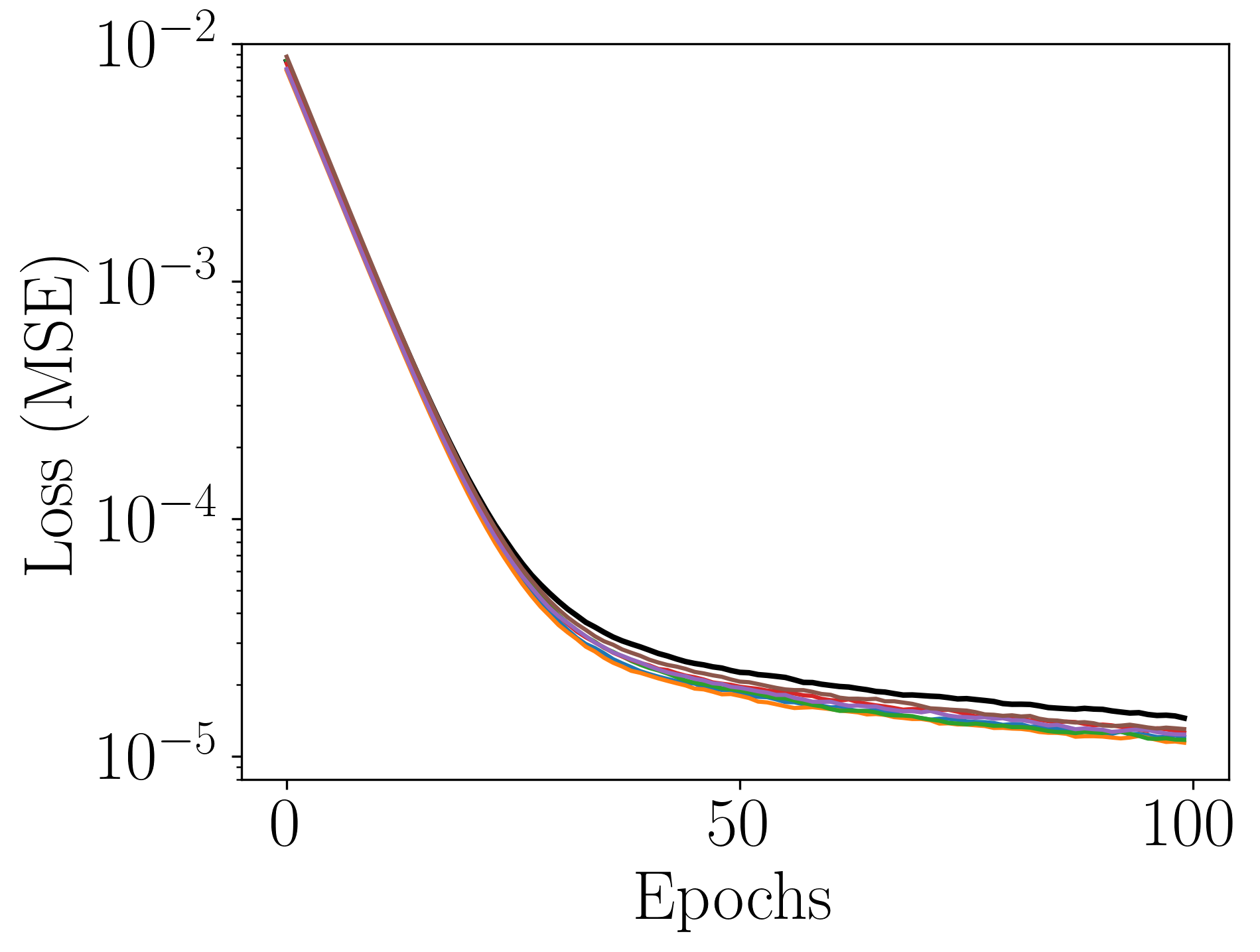}
        \caption{Training loss}
    \end{subfigure}
    \begin{subfigure}[b]{0.475\linewidth}
        \includegraphics[width=\linewidth]{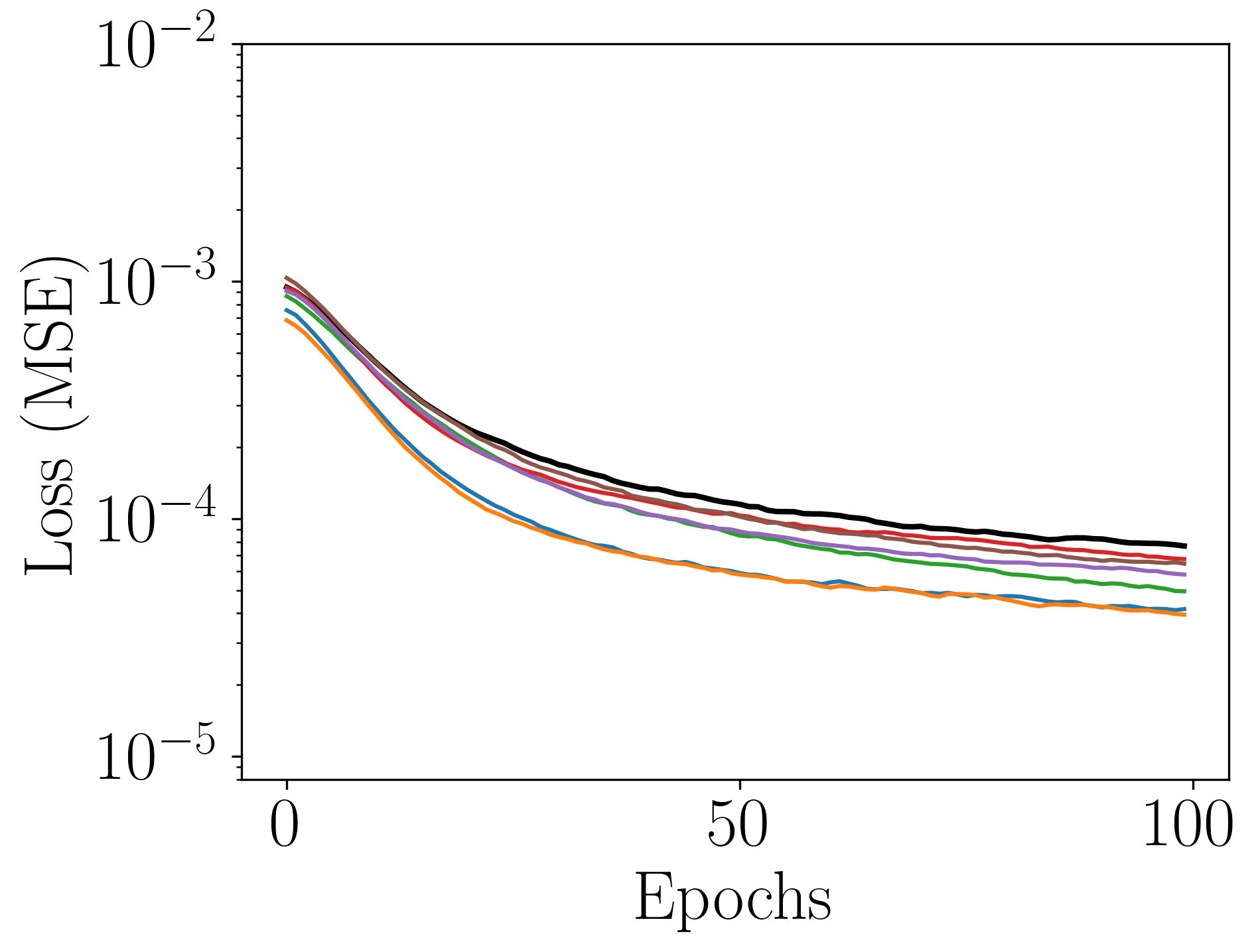}
        \caption{Validation loss}
    \end{subfigure}
    \caption{Comparison of the training and validation loss for the Henon Heiles system with different injection configurations.}
    \label{fig:henonheiles_training_loss_comparison}
\end{figure}

\begin{figure}
    \centering
    \begin{adjustbox}{max width=0.5\linewidth}
        \begin{tikzpicture}
            \begin{customlegend}[legend columns=4,legend style={draw=none, column sep=1ex},legend entries={$x$,$y$,$\dot{x}$,$\dot{y}$, Truth,Prediction,95\% conf.}]
                \addlegendimage{Tblue,sharp plot}
                \addlegendimage{Tred,sharp plot}
                \addlegendimage{Tgreen,sharp plot}
                \addlegendimage{Torange,sharp plot}
                \addlegendimage{black, sharp plot}
                \addlegendimage{black, dashed,sharp plot}
            \addlegendimage{black!20, fill=black!20, area legend}
            \end{customlegend}
        \end{tikzpicture}
    \end{adjustbox}

    \begin{subfigure}[b]{0.475\linewidth}
        {\includegraphics[width=\linewidth]{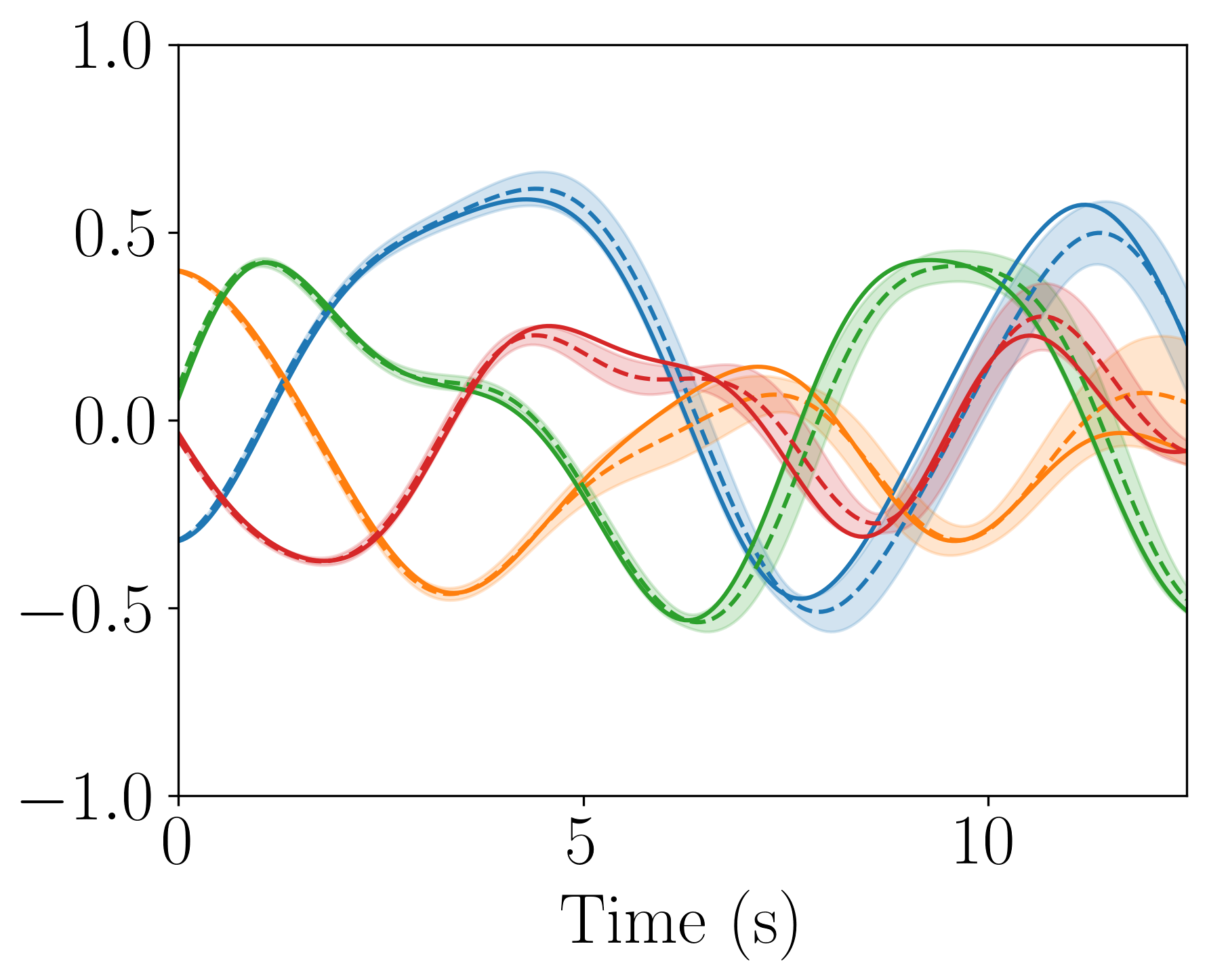}}
        \caption{No injection}
    \end{subfigure}
    \begin{subfigure}[b]{0.475\linewidth}
        \includegraphics[width=\linewidth]{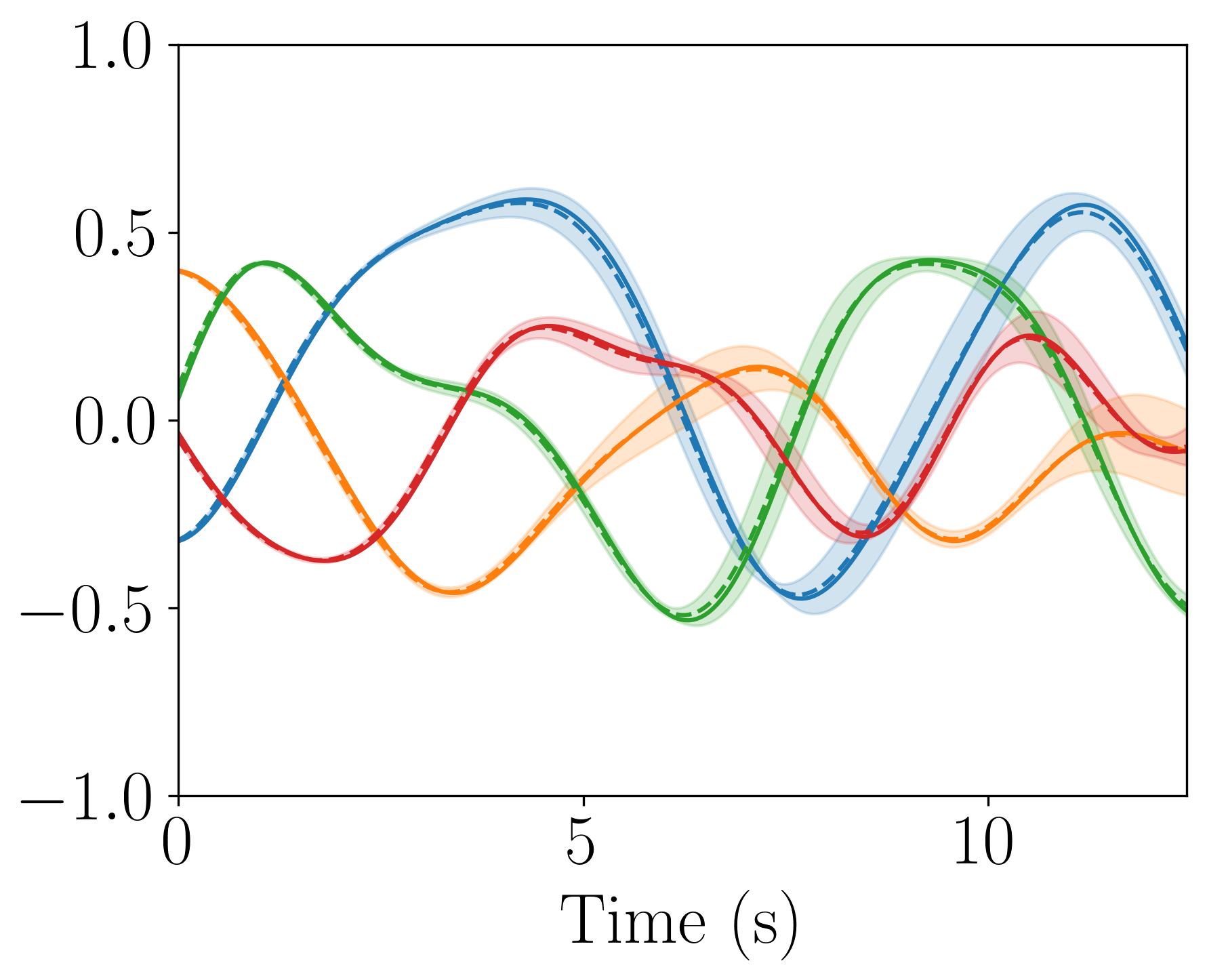}
        \caption{Injection $xy$ in the first layer}
    \end{subfigure}
    \begin{subfigure}[b]{0.475\linewidth}
        \includegraphics[width=\linewidth]{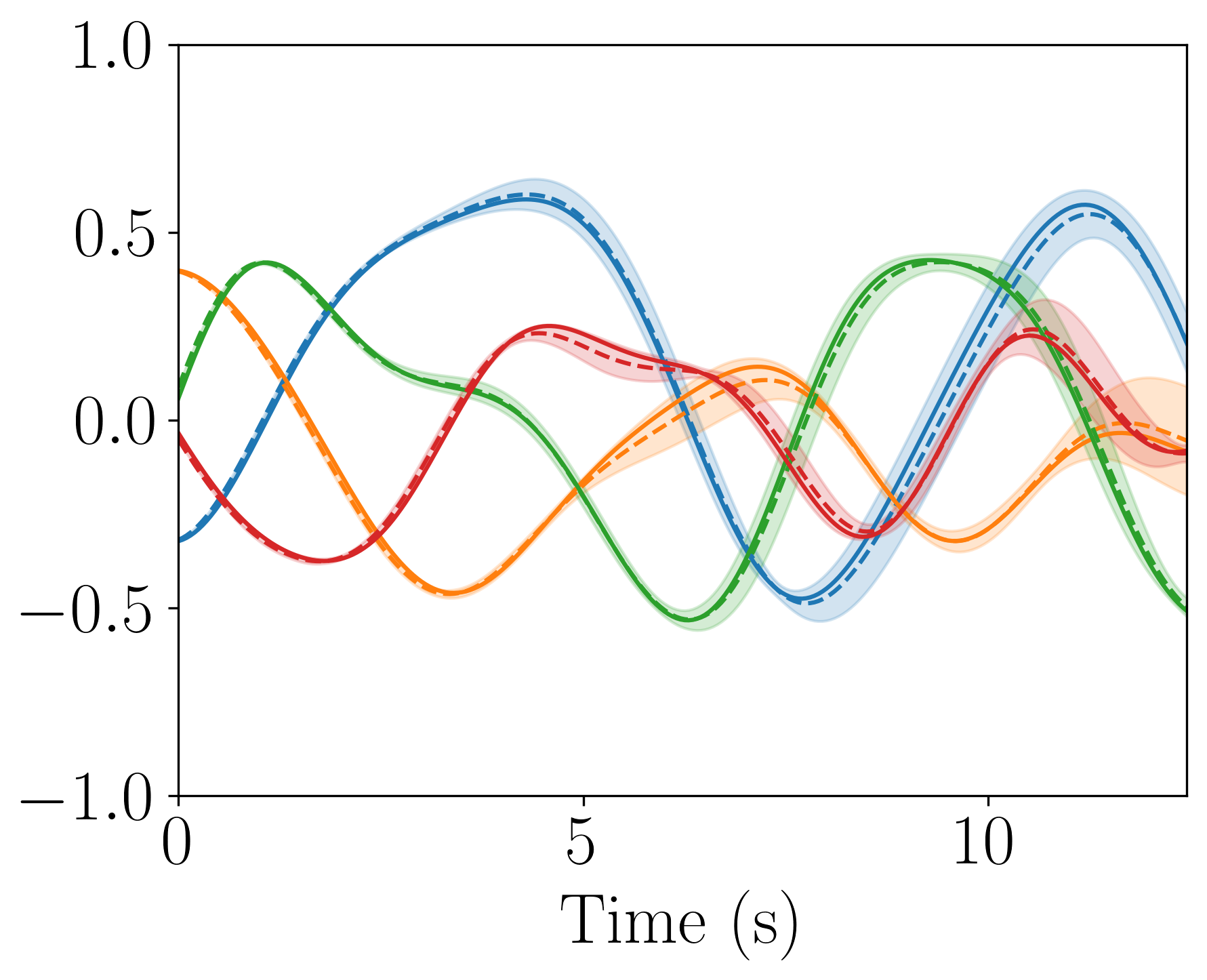}
        \caption{Injection $xy$ in second layer}
    \end{subfigure}
    \begin{subfigure}[b]{0.475\linewidth}
        {\includegraphics[width=\linewidth]{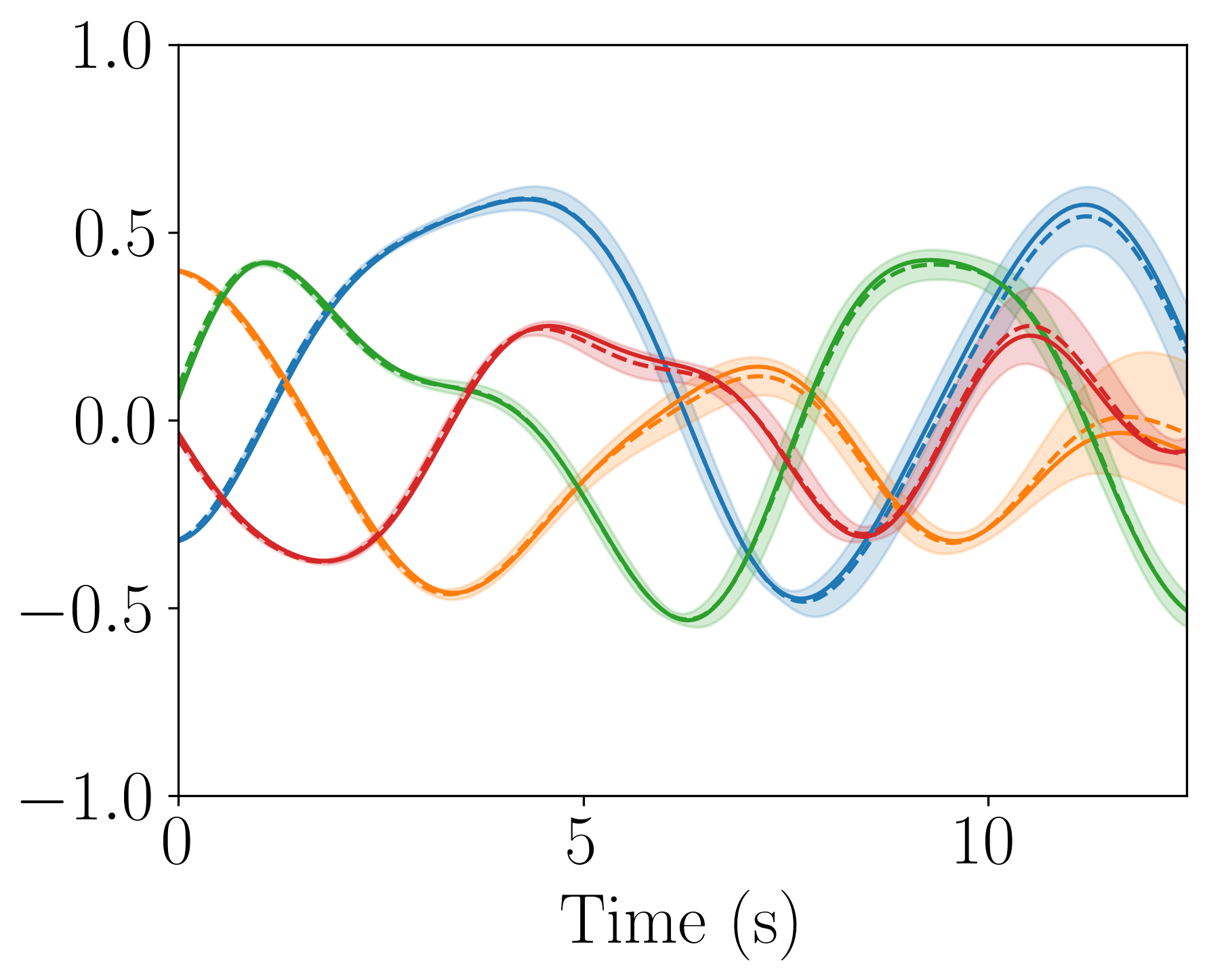}}
        \caption{Injection $xy$ in third layer}
    \end{subfigure}
    
    \caption{Rolling forecast for the Henon-Heiles system without and with $xy$ injection at different layers. Injecting functions appears to significantly aid the learning of this system. The best results are achieved by into the last hidden layer.}
    \label{fig:henonheiles_reg_forecast}
\end{figure}

For this system, all models performed very similarly, as can be seen in Figure \ref{fig:henonheiles_reg_forecast} for the $xy$ injection. Figure \ref{fig:henonheiles_training_loss_comparison} shows that the injected models converge slightly faster and reach an overall lower validation loss.

\section{Conclusion and future work}
\label{sec:conclusionandfuturework}
The physics-guided neural network (PGNN) framework was applied to a set of five distinct dynamical systems represented by first and second order non-linear ordinary differential equations. Three of the systems had two suitable injection terms which were also investigated, for a total of 8 combinations of system and injection terms. All possible injection layers were evaluated and compared. The main conclusions from the work are as follows:

\begin{itemize}
    \item Knowledge injection can accelerate training and lead to better convergence. However, knowledge injection does not guarantee an improvement in performance.
    \item Accuracy of the models can in general be improved through knowledge injection. For the Van der Pol system, knowledge injection helped the models capture the fast transients, which were relatively underrepresented in the dataset. However, for some systems like Henon-Heiles system the improvements were not significant.
    \item The study remains inconclusive regarding the impact of knowledge injection on model uncertainty. For the Duffing oscillator we see a shrinkage in the uncertainty but for Van der Pol, we see an increase. However, we should stress that lower uncertainty with poor prediction would not be desirable.
\end{itemize}

The first limitation of this work is that the choice of injection layer was found to have a significant impact on performance, and the results do not show how such a choice could be made a priori. A second limitation is that we do not address how to identify the correct information to inject into the hidden layers of the network. We simply tried all combinations of injection layers and terms, but it is clear that this will scale poorly for network architectures with more hidden layers or multiple injection terms. Hyperparameter search algorithms (for example genetic algorithms due to the discrete optimisation variable) may be helpful tools to choose an effective injection layer. However, this still involves training multiple models. A simpler solution to both problems is to make all injection features available to all layers via skip connections, although preliminary results in this direction have shown that this does not reach the same level of performance as the best single injection layer. Combining this approach with sparsifying regularisation may bias the network towards selecting the best injection layer and term. Testing the PGNN approach with deeper architectures and recurrent NNs could also elucidate the role of the injection layer, and should be investigated. Another method that could help identify suitable injection terms is symbolic regression based on gene expression programming (GEP). By running a large SR ensemble on the data and selecting the most frequently appearing terms, it might be possible to collect useful injection terms. 

We see extensions of PGNNs as being useful for modelling more complex systems with rich dynamics and interactions with the environment. Significant assumptions are typically made about the nature of environmental forces and practitioners often defer to the data, e.g. when modelling the average wind force on marine vessels \citep{fossen_marine_2021,isherwood_wind_1972,blendermann_parameter_1994}. There is already much work where more advanced ML techniques such as reinforcement learning (RL) are applied to these systems \citep{meyer2020taming}. The improved training characteristics and low overhead of PGNNs may prove useful in these RL contexts, where data efficiency is typically quite poor. Furthermore, the learned weighting of explicit features arguably makes the models more interpretable, which is often seen as desirable in safety-critical contexts, although more work is needed in this direction. PGNNs may also require fewer parameters than conventional DNNs to model the same data, which could enable the use of existing robustness verification algorithms \citep{liu_algorithms_2019}, vastly improving confidence in these systems during deployment.

The attractiveness of prior knowledge injection is that it generalises two of the most common hybridisation methods: input feature engineering and output error correction. By injecting arbitrary features into the intermediate layers of a neural network, we begin to open the proverbial black box and recognise its potential as a general-purpose feature-learner and feature-selector, powered by stochastic optimisation.

\section*{Acknowledgments}
A.R. is grateful for the support received by the Research Council of Norway and the industrial partners of the following projects: EXAIGON--{\em Explainable AI systems for gradual industry adoption\/} (grant no. 304843). O.S. gratefully acknowledges the U.S. DOE Early Career Research Program support (DE-SC0019290), and the National Science Foundation support (DMS-2012255).

\bibliographystyle{elsarticle-harv}
\bibliography{references,library}
\end{document}